\documentclass[10pt,twocolumn,letterpaper]{article}

\usepackage[pagenumbers]{cvpr} 

\usepackage{amstext}
\usepackage{amsmath}
\usepackage{array}
\usepackage{subcaption}
\usepackage{threeparttable}
\usepackage{makecell}
\usepackage{booktabs}
\usepackage{stfloats}
\usepackage{multirow}
\usepackage{bbding}
\usepackage{dsfont}
\usepackage{rotating}
\usepackage{tabularx}
\usepackage[table]{xcolor}
\usepackage{colortbl}
\usepackage{wrapfig}
\usepackage{setspace}
\usepackage{algorithmic}
\usepackage[ruled,vlined,linesnumbered]{algorithm2e}  
\usepackage{graphicx}
\usepackage{pifont}
\usepackage{enumitem}

\usepackage[normalem]{ulem}

\usepackage[most]{tcolorbox}
\definecolor{best}{RGB}{255,232,204}    
\definecolor{second}{RGB}{221,235,247}  

\definecolor{LG}{RGB}{198,239,206} 
\definecolor{MG}{RGB}{149,215,165} 
\definecolor{DG}{RGB}{99,190,123}  
\definecolor{DDG}{RGB}{49,165,90}  
\definecolor{LR}{RGB}{255,199,206} 
\definecolor{MR}{RGB}{251,152,156} 
\definecolor{DR}{RGB}{248,105,107} 
\definecolor{DDR}{RGB}{244,58,58} 

\definecolor{RAcolor}{HTML}{FF8000} 
\definecolor{RBcolor}{HTML}{00ADEE} 
\definecolor{RCcolor}{HTML}{EC0090} 

\newcommand{\Logo}{\raisebox{-0.35\height}{\includegraphics[width=2em]{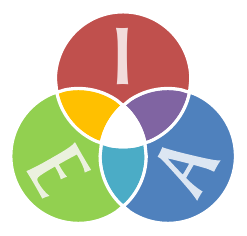}}}
\newcommand{\agent}{\textsc{IEA}}

\usepackage{ifthen}
\usepackage{ulem}
\usepackage{CJKutf8}

\newif\ifshowcomments
\showcommentsfalse

\newcommand{\CL}[2][]{
\begin{CJK}{UTF8}{gbsn}\ifshowcomments
  \ifthenelse{\equal{#1}{done}} 
    {\textcolor{blue}{(\sout{<Solved> Lu}: #2)}}
    {\ifthenelse{\equal{#1}{key}}
        {\textcolor{blue}{\textbf{(Lu: #2)}}} 
        {\textcolor{blue}{(Lu: #2)}}}
        \fi\end{CJK}}

\newcommand{\ZZC}[2][]{
\begin{CJK}{UTF8}{gbsn}\ifshowcomments
  \ifthenelse{\equal{#1}{done}} 
    {\textcolor[HTML]{5e9832}{(\sout{<Solved> Zichen}: #2)}}
    {\ifthenelse{\equal{#1}{key}}
        {\textcolor[HTML]{5e9832}{\textbf{(Zichen: #2)}}} 
        {\textcolor[HTML]{5e9832}{(Zichen: #2)}}}
        \fi\end{CJK}}

\newcommand{\ZDY}[2][]{
\begin{CJK}{UTF8}{gbsn}\ifshowcomments
  \ifthenelse{\equal{#1}{done}} 
    {\textcolor[HTML]{242671}{(\sout{<Solved> ZDY}: #2)}}
    {\ifthenelse{\equal{#1}{key}}
        {\textcolor[HTML]{242671}{\textbf{(ZDY: #2)}}} 
        {\textcolor[HTML]{242671}{(ZDY: #2)}}}
        \fi\end{CJK}}
        
\definecolor{cvprblue}{rgb}{0.21,0.49,0.74}
\usepackage[pagebackref,breaklinks,colorlinks,allcolors=cvprblue]{hyperref}

\usepackage{float}

\usepackage[capitalize]{cleveref}
\usepackage[skip=2pt]{caption}

\def\paperID{37520} 
\def\confName{CVPR}
\def\confYear{2026}

\title{\Logo{} Amateur-Friendly Conversational Image Editing Agent via Three Stages of Multitask Alignment}

\author{Zichen Zhu$^{14*\ddagger}$, Yuheng Sun$^{1*}$, Mingxuan Zhu$^{1*}$, Wenjie Ma$^1$, Situo Zhang$^1$\\ Zhexiang Wang$^1$, Ziyue Yang$^1$, Danyang Zhang$^1$, Kunyao Lan$^1$, Zihan Zhao$^1$\\ Dingye Liu$^1$, Siqi Xiang$^{3\dagger}$, Lu Chen$^{125\dagger}$, Kai Yu$^{15\dagger}$ \\
\normalsize$^1$ \textit{X-LANCE Lab, School of Computer Science, Shanghai Jiao Tong University} \\
\normalsize$^2$ \textit{Shanghai Innovation Institution} \quad 
\normalsize$^3$ \textit{Huawei Technologies Ltd.} \\
\normalsize$^4$ \textit{Nanyang Technological University} \quad 
\normalsize$^5$ \textit{Jiangsu Key Lab of Language Computing}\\
 {\tt\normalsize \{JamesZhutheThird, chenlusz, kai.yu\}@sjtu.edu.cn, xiangsiqi@huawei.com}
}

\begin{document}

\maketitle

\begingroup
\renewcommand\thefootnote{}
\footnotetext{
$^{*}$ Equal contribution 
$^{\ddagger}$ Project lead 
$^{\dagger}$ Corresponding authors 
}
\endgroup

\begin{abstract}
Current image editing software often hinges on fixed filters or expert tuning, leaving a gap between amateur users’ intent and outcomes. Creations by generative models may contain artifacts, implausible details, or stylistic drift away from photorealism and offer little insight into why an edit was made.  
We propose IEA, a conversational \textbf{I}mage \textbf{E}diting \textbf{A}gent that learns to operate parameterized tools in an explicit, interpretable action space. IEA is trained via a three-stage multitask pipeline: (1) SFT on distilled expert edits, (2) GRPO with rewards for likeness improvement, tool usefulness, and intent summarization, and (3) large-scale synthetic fine-tuning to jointly master image editing, refinement, and user intent summarization. By manipulating 16 editing tools step by step, IEA produces transparent edit traces that can be inspected and debugged. In quantitative experiments, it attains a lower pixel distance on the edit task and a higher ROUGE-L on the summary task than strong baselines. In user studies, it ranks best among tool-calling methods for instruction following while surpassing generative methods in overall perceptual quality. Our results validate interpretable, tool-centric VLMs as a reliable path to human instruction-guided image retouching. Our data and code are released at
\href{https://github.com/OpenDFM/Image_Edit_Agent}{this link}.
\end{abstract}    
\vspace{-0.3cm}
\section{Introduction}
\label{sec:intro}
The proliferation of digital photography has made image editing a common activity for both casual users and professionals. However, existing editing tools exhibit a clear polarization: professional desktop software (such as Photoshop and Lightroom) offers powerful functions but has a steep learning curve that deters amateur users; meanwhile, mobile-based ``one-click filters'' or automatic enhancement features, while convenient, lack fine-grained control and personalized expression capabilities. These limitations often prevent current tools from accurately meeting users' unique subjective aesthetic intentions. Users increasingly expect to obtain satisfactory editing results through simple natural language descriptions (e.g., ``Give me a warm and comfortable atmosphere'' or ``Darken the background slightly to make the human face more prominent''), which represents a significant pain point that existing tools fail to address effectively. 

\begin{figure}[t]
  \includegraphics[width=\linewidth]{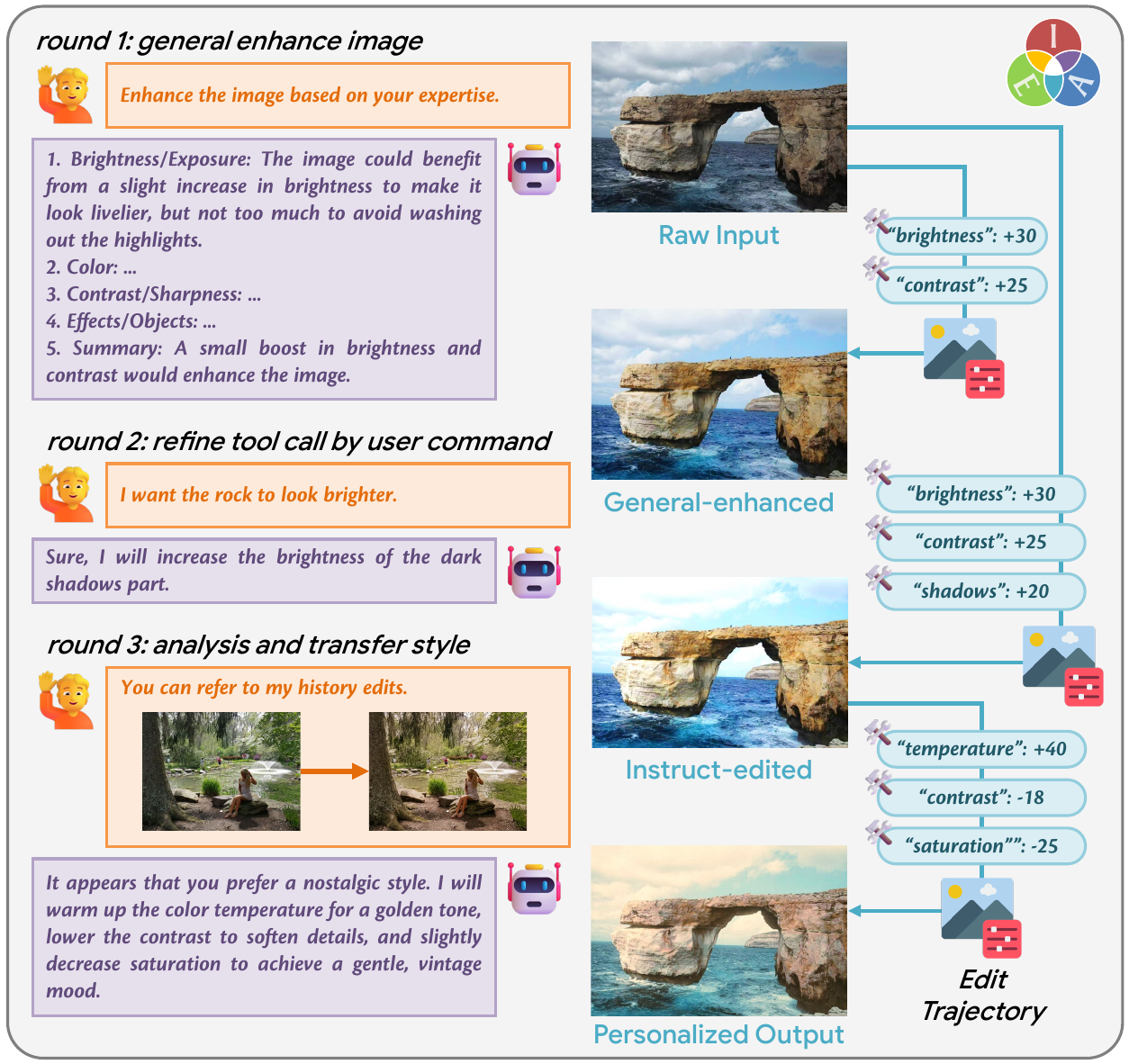}
  \caption{User can easily interact with \agent{} to edit the image in the styles they prefer, or simply refine in general expertise. \agent{} can also learn user intents based on their previous attempts.}
  \label{fig:teaser}
\end{figure}

With the development of Variational Autoencoders (VAEs)~\cite{kingma2013auto}, diffusion techniques~\cite{rombach2021highresolution,reuss2024multimodal}, and autoregressive models, image generation systems~\cite{ramesh2022hierarchical,wu2025qwen,gpt-image,brooks2023instructpix2pix,zhang2023magicbrush} can create images from scratch with personalized instructions or redraw based on references. However, these pipelines can be compute-intensive, and the generated images may introduce artifacts, implausible details, or stylistic drift away from photorealism. 

Many VLM-empowered agents have made promising contributions to tool calling and GUI control~\cite{sun-etal-2022-meta,zhang2024ufo,zhu-etal-2025-moba}. Meanwhile, frontier visual language models (VLMs) such as GPT-4.1~\cite{gpt4.1}, Gemini-2.5-Pro~\cite{gemini2.5}, and Qwen2.5-VL~\cite{Qwen2.5-VL} demonstrate strong abilities to understand image content and user instructions and to provide editing guidance. A practical direction is to teach VLMs to operate image editing software directly. Recent systems~\cite{lin2025jarvisart,chen2025photoartagent} take a user instruction and an image as input, and generate an Adobe Lightroom style template as output; the user then applies the template to obtain the final image. However, the external editor introduces time-consuming communication to produce a result, and reward signals are often tied only to overall visual effects, which may encourage redundant or suboptimal tool usage.

To bridge this gap and improve training efficiency, we introduce \agent{}, an amateur-friendly conversational \textbf{I}mage \textbf{E}diting \textbf{A}gent trained via three stages of human preference alignment. As shown in Figure~\ref{fig:teaser}, \agent{} allows users to interact through natural language commands, automatically generating specific and interpretable tool calls to achieve non-destructive, high-fidelity photo editing. This approach preserves the authenticity of editing results while enabling seamless integration into existing editing workflows.

Our training framework comprises three core stages. The Supervised Fine-Tuning (SFT) stage constructs a high-quality dataset by distilling tool-parameter plans from expert edits using capable VLMs and a lightweight heuristic parameter search to better match expert outcomes. The Image-Edit task requires the model to analyze the image along with user instructions and provide tool calls and reasoning thoughts. We also introduce a reversed Image-Summary task, aiming to summarize user instructions based on editing results. The Reinforcement Learning (RL) stage employs Group Relative Policy Optimization (GRPO) to refine the agent's policy with rewards that capture both distance improvement to reference edits and the marginal usefulness of each tool invocation. We also trained a small reward model to guide the reversed image summary task. Finally, the enhancement stage leverages large-scale synthetic supervision to broaden instruction coverage and, crucially, to enable Image-Refine ability, incrementally improving an earlier attempt based on user feedback.

\agent{} surpasses all baseline models in quantitative experiments, achieving an average pixel distance of $0.1034$ to the reference image on Image-Edit and a Rouge-L score of $0.2575$ on Image-Summary on the expert test dataset. We also conducted a user study where \agent{} ranked highest among tool-calling methods for instruction following and outperformed all baselines in overall image quality assessment. These results suggest that combining VLMs with symbolic, controllable tools is an effective pathway toward reliable, interpretable, and preference-aligned image editing. Our contributions are listed below:
\begin{itemize}
    \item We present \agent{}, a VLM-driven image editing agent that analyzes the given image and instruction to select appropriate editing tools and parameters, summarizes user intents from past edits, and refines results in response to feedback.
    \item We construct a high-quality dataset with $\sim$29k samples distilled from GIER, and a large-scale synthetic dataset with $\sim$400k samples. 
    \item We develop a three-stage training recipe: (i) SFT to initialize tool usage and instruction summarization; (ii) GRPO with likeness-improvement and usefulness rewards to align parameters with perceptual goals, plus an alignment reward for the summary task; and (iii) synthetic data to enable image refinement and improve reasoning, tool proficiency, and generalization.
    \item Quantitative experiments and the user study demonstrate that \agent{} delivers impressive instruction following and image quality compared with both diffusion-based generation and tool-calling baselines.
\end{itemize}

\section{Related Work}
\label{sec:related}
\subsection{Automated Image Retouching}

Automated image retouching has evolved through three major paradigms.
\textbf{(1) Interpretable pipelines.} Early methods relied on sequential, parameterized pipelines with differentiable filters or RL policies for global adjustments~\cite{Lee_2024_CVPR,Ke_2023_CVPR,Shi_2022_CVPR,Conde_2025_ICCV}. While efficient and traceable~\cite{park2018distort,hu2018exposure,shi2020benchmark}, these approaches were limited in expressiveness and provided only coarse global control.
\textbf{(2) Diffusion-based and instruction-following editing.}
Diffusion models enabled high-fidelity transformations through conditional denoising~\cite{meng2021sdedit,lugmayr2022repaint,hertz2022prompt,mokady2023null,DBLP:conf/cvpr/KawarZLTCDMI23,Avrahami_2022_CVPR,Yang_2023_CVPR}, and instruction-following diffusion extended editing to text-driven image manipulation~\cite{ramesh2022hierarchical,wu2025qwen,gpt-image,brooks2023instructpix2pix,zhang2023magicbrush,NEURIPS2024_05a30a0f,Conde_2025_ICCV}.
More recent work improves efficiency and controllability via fast or inversion-free sampling and MLLM-guided controllers~\cite{deutch2024turboedittextbasedimageediting,Xu_2024_CVPR,Huang_2024_CVPR,fu2024guiding}, and explores action- and reasoning-centric editing from videos and simulation~\cite{krojer2024aurora}.
However, these methods still suffer from destructive regeneration, limited local attribute control, and substantial computational overhead. Even unified VLMs continue to struggle with these challenges.
\textbf{(3) VLM-based tool-calling and GUI agents.}
The latest paradigm performs non-destructive, parameterized retouching by using VLMs to call professional editing tools.
These models learn tool--parameter relationships to generate executable editing plans, enabling interpretability and resolution-agnostic operation~\cite{lin2025jarvisart,chen2025photoartagent}.
In parallel, GUI agents~\cite{zhang2024ufo,zhu-etal-2025-moba,liu2024agentbench} attempt to operate real interfaces directly, but face difficulties in long-horizon interaction, UI variability, and sparse reward signals that hinder robust policy learning.
Our approach integrates multimodal understanding with expert editing tools through a collaborative agent-human loop, enabling both efficient and precise refinements for high-quality retouching.

\begin{figure*}[t]
    \centering
    \includegraphics[width=\linewidth]{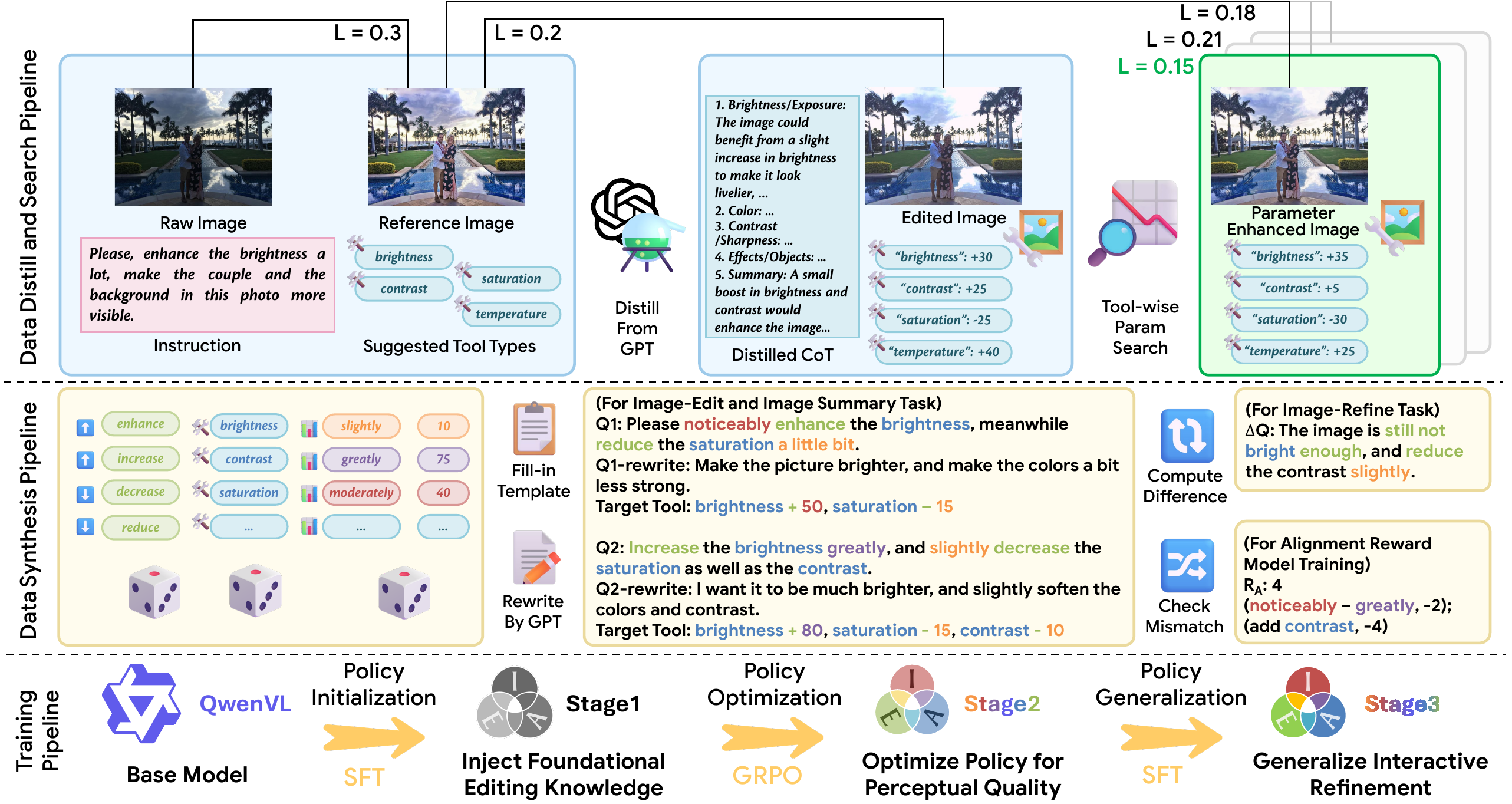}
    \caption{
        Overview of the \agent{} framework: Data Pipelines and Three-Stage Training Process.
    }
    \label{fig:pipeline}
\vspace{-0.5cm}
\end{figure*}

\subsection{Learning to Use Tools}
Large Language Models (LLMs) and Visual Language Models (VLMs) have demonstrated remarkable cross-domain generalization capabilities~\cite{10.5555/3600270.3602281}, enabling them to tackle diverse tasks such as natural language understanding and reasoning. Nevertheless, despite their versatility, LLMs exhibit inherent limitations. They are prone to generating hallucinations~\cite{10.1145/3571730,xu2025reducing,xu2025delusionslargelanguagemodels}, lack access to dynamic and real-time information sources~\cite{10.1145/3637528.3671470}, and face challenges in adapting to user-specific personalization requirements~\cite{zhang2025personalizationlargelanguagemodels}. To mitigate these challenges, researchers have increasingly explored equipping LLMs with tool-use capabilities, which complement model-internal knowledge with external functionalities. Such integration not only reduces hallucinations by grounding responses in reliable resources but also extends the applicability of LLMs to time-sensitive and personalized contexts.

The literature has broadly categorized the tool-use abilities of LLMs into five interconnected dimensions. \textbf{(1) Task Planning} can be divided into tuning-free methods~\cite{yao2023react,10.5555/3666122.3667779,DBLP:conf/iclr/ZhuangC0MBRS024,10.1007/978-3-031-73254-6_6,qu2025from,zhu-etal-2025-moba,lan2024depressiondiagnosisdialoguesimulation}, 
which generally yield improvements across diverse scenarios without additional training, 
and tuning-based methods~\cite{schick2023toolformer,qian-etal-2024-toolink,gao-etal-2025-efficient,10.5555/3737916.3739641}, 
which tend to achieve more substantial gains in domain-specific settings. 
\textbf{(2) Tool Selection} has been studied in two major paradigms: 
retriever-based selection~\cite{lei-etal-2023-unsupervised,DBLP:conf/iclr/YuanC000J24,zheng-etal-2024-toolrerank,10.1145/3627673.3679847} 
and LLM-based selection~\cite{qin2024toolllm,DBLP:journals/corr/abs-2403-00839,mekala-etal-2024-toolverifier}. 
\textbf{(3) Tool Calling} focuses on accurately invoking external functions or APIs~\cite{hao2023toolkengpt,xu2025alignmentefficienttoolcalling}. 
\textbf{(4) Response Generation} concerns incorporating tool outputs into coherent and contextually appropriate responses~\cite{shi-etal-2024-learning}. 
\textbf{(5) Agentic and environment-grounded tool use} examines how LLMs interact with rich visual, API, browser, robotic, or multienvironment interfaces~\cite{wu2023visualchatgpttalkingdrawing,zeng2023socratic,Suris_2023_ICCV,nakano2022webgptbrowserassistedquestionansweringhuman,patil2024gorilla,DBLP:conf/emnlp/LiZ000YLHL23,saycan2022arxiv,liu2024agentbench,wang2023voyager,lu2023chameleon}, 
highlighting the challenges of long-horizon decision-making, environmental variability, and robust multistep execution.

Overall, enabling LLMs to flexibly acquire tool-use abilities—whether through tuning-free or tuning-based approaches—constitutes a crucial direction for future research. Such capabilities hold promise for advancing the reliability, real-time adaptability, and customization potential of large models.

\vspace{-0.2cm}
\section{Methodology}
We illustrate the overall pipeline in \Cref{fig:pipeline}. In this section, we first introduce the simulation editor for real-time image editing in \Cref{sec:env}. Since the training process of \agent{} contains 3 stages, we will correspondingly introduce: \textbf{Stage 1}: Policy Initialization (\cref{sec:stage1}), \textbf{Stage 2}: Policy Optimization (\cref{sec:stage2}), \textbf{Stage 3}: Policy Generalization (\cref{sec:stage3}).

\subsection{Image Editing Simulation Environment}
\label{sec:env}

The foundation of our work is a simulated image editing environment that mimics the functionality of standard image editing software like Adobe Lightroom. To provide real-time editing results during training for reward calculation and feedback, we implemented the environment using \texttt{matplotlib}, referencing commonly available tools in modern image editing applications. The built-in editor can be easily deployed on a local Ubuntu server and processes each image in 50 to 300 ms. It supports 16 adjustment parameters, each of which can be adjusted in integer steps from $-100$ to $100$. 
The complete list of these tools is provided in the supplementary material. 

\subsection{Stage 1: Policy Initialization}
\label{sec:stage1}

\paragraph{Objective.} We initialize the policy with supervised fine-tuning (SFT) on carefully crafted training data so that the model learns (i) the input–output format of the tasks; and (ii) reasonable prior knowledge over tool selection and parameter magnitudes before any reinforcement learning. In this stage, we introduce two dual tasks: \emph{Image-Edit} (predict tool calls and parameters given an image with/out specific instruction) and \emph{Image-Summary} (summarize user intents and predict instruction given a before/after pair).

\paragraph{Semi-Supervised Data Collection}

We begin with the GIER dataset~\cite{gier}, which comprises Internet photos paired with expert human retouching. The annotations include revised user instructions as well as predictions of the editing tools applied during the retouching process. Since the built-in editor in this work focuses on global adjustments (e.g., exposure, contrast, saturation, tint, temperature, and sharpness), we excluded samples involving local editing tools, resulting in a final set of 2.4k image pairs.

Since GIER specifies only tool types without precise numeric parameters, we leverage a strong VLM, \texttt{GPT-4.1}~\cite{gpt4.1}, to infer the missing details and reasoning steps. For each filtered pair, we provide: (1) a complete list of tools with parameter explanations (see the supplementary material) from our built-in editor; (2) both the original image and its expert-retouched counterpart; and (3) rewritten human instructions along with the annotator-predicted tool types. We prompt the VLM to generate both: (1) step-by-step chain-of-thought reasoning; and (2) a sequence of tool calls with explicit parameters. The tool calls follow a JSON format compatible with our built-in editor. To further reduce the sim-to-real gap, we perform a lightweight heuristic search around the distilled parameters—using coarse-to-fine line search per dimension with early stopping—to more closely match expert retouching results (pseudocode provided in the supplementary material).

\paragraph{Image-Edit Task Formulation.} 
For Image-Edit task, given a user instruction query $Q$ and an original image $I_{\text{ori}}$, the model $\mathcal{M}$ predicts a set of tool parameters $T={t_1,\dots,t_m}$:
\vspace{-0.2cm}
\begin{equation}
\mathcal{M}(I_{\text{ori}}, Q) \rightarrow T.
\end{equation}
\begin{figure}[h]
\vspace{-0.5cm}
    \centering
    \includegraphics[width=\linewidth]{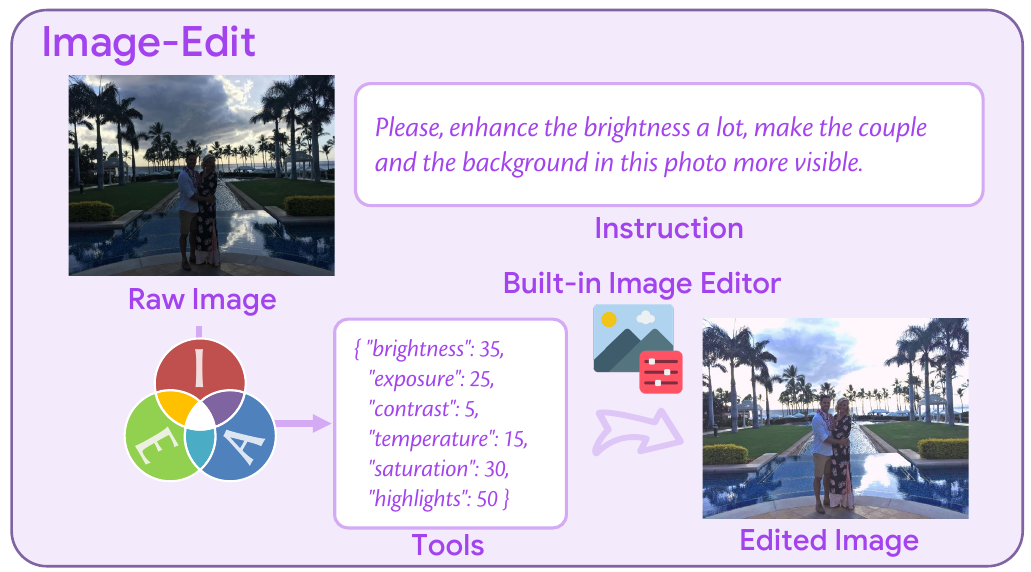}
    \caption{Example of Image-Edit task.}
    \label{fig:image-edit}
\vspace{-0.8cm}
\end{figure}

\paragraph{Image-Summary Task Formulation.} 
For the task Image-Summary, given the original image $I_{\text{ori}}$ and an edited historical image $I_{\text{his}}$, the model generates a concise instruction $Q$ that encapsulates the user’s editing preferences and intents:
\vspace{-0.2cm}
\begin{equation}
\mathcal{M}(I_{\text{ori}}, I_{\text{his}}) \rightarrow Q.
\end{equation}
\begin{figure}[h]
\vspace{-0.7cm}
    \centering
    \includegraphics[width=\linewidth]{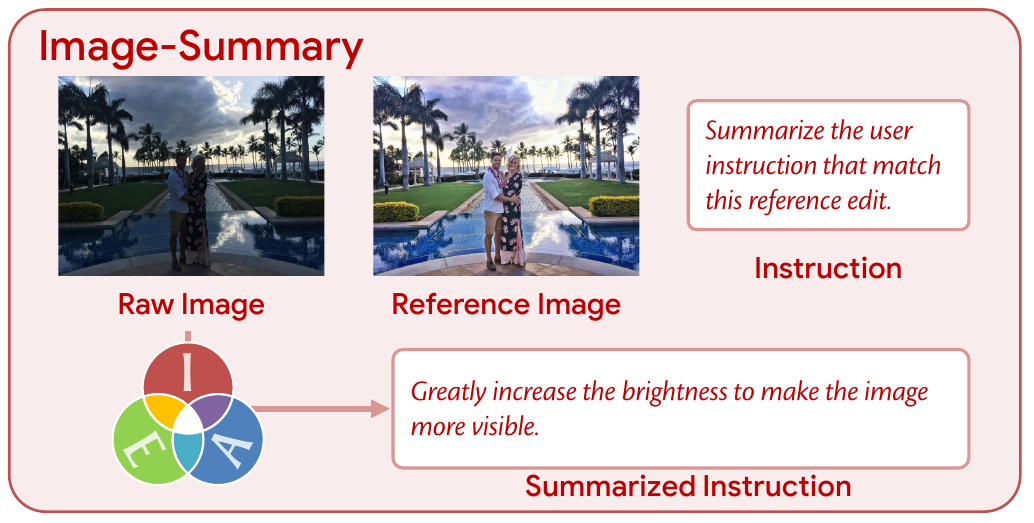}
    \caption{Example of Image-Summary Task.}
    \label{fig:image-summary}
\vspace{-0.5cm}
\end{figure}

\subsection{Stage 2: Policy Optimization}
\label{sec:stage2}

\paragraph{Objective.} After SFT, we optimize the policy using Group Relative Policy Optimization (GRPO) \cite{shao2024deepseekmath}, sampling $K$ rollouts per prompt and updating the policy based on our carefully designed rewards: (1) $R_L$ encourages the retouched image to closely resemble expert-level retouching; (2) $R_U$ penalizes redundant or ineffective tool calls; and (3) $R_A$ ensures that summarized instructions can vary in expression while preserving the same intent.

\paragraph{Image-Edit Reward.} Let $\mathcal{E}(I_{\text{ori}},T)$ denote the image produced by our built-in editor $\mathcal{E}$ when applying the set of tool parameters $T$ to $I_{\text{ori}}$. Let $\mathcal{L}(\cdot,\cdot)$ be the average of L1 and L2 distances between two images. We define a normalized improvement term, referred to as the likeness improvement reward: 
\vspace{-0.2cm}
\begin{equation}
R_L = \max\!\Big(-1,\frac{\mathcal{L}(I_{\text{ori}}, I_{\text{ref}})-\mathcal{L}(\mathcal{E}(I_{\text{ori}}, T), I_{\text{ref}})}{\mathcal{L}(I_{\text{ori}}, I_{\text{ref}})}\Big)\in(-1,1]
\end{equation}
which quantifies how much closer the edited image is to the reference image $I_{\text{ref}}$ relative to the original.
To encourage efficient and purposeful use of tools, we introduce a usefulness reward based on marginal utility:
\vspace{-0.2cm}
\begin{equation}
\begin{aligned}
R_U = \frac{1}{|T|} \sum_{t\in T} 
\mathds{1}\Big(
    \mathcal{L}(\mathcal{E}(I_{\text{ori}},T\setminus\{t\}), I_{\text{ref}})  \\[-10pt]
    > \mathcal{L}(\mathcal{E}(I_{\text{ori}},T), I_{\text{ref}})
\Big)
\end{aligned}
\end{equation}
\vspace{-0.5cm}

\begin{figure}[h]
    \centering
    \includegraphics[width=\linewidth]{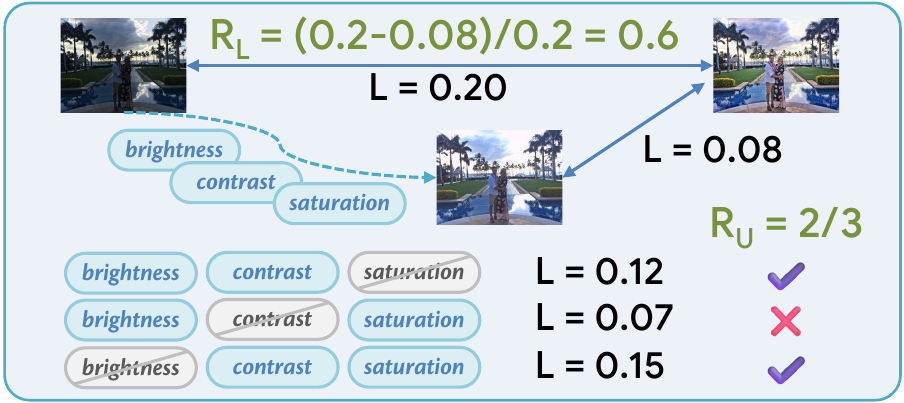}
    \caption{Illustration of likeness improvement reward $R_L$ and usefulness reward $R_U$.}
    \label{fig:reward}
\end{figure}

\noindent That is, if removing a particular tool parameter $t$ causes an increase in distance from $I_{\text{ref}}$, then that tool is considered truly beneficial. The illustration of two edit reward is shown in \Cref{fig:reward}. 

We weigh the two rewards to obtain the final reward for task Image-Edit:
\vspace{-0.2cm}
\begin{equation}
R_{\text{edit}} = \alpha R_L + (1{-}\alpha) R_U
\end{equation}
\vspace{-0.5cm}

\noindent 
This formulation explicitly associates reward with perceptual similarity while penalizing unnecessary or ineffective tool usage.

\paragraph{Image-Summary Reward.} We train a lightweight reward model (RM) to evaluate the agreement between a reference instruction $Q_{\text{ref}}$ and a predicted instruction $Q_{\text{pred}}$. The RM produces an alignment reward $R_A \in [-10,10]$, reflecting both semantic consistency and specificity. Instructions with exact semantic matches are assigned scores near $+10$. Outputs that are partially aligned but omit some key attributes receive modest positive scores, whereas incorrect key information results in negative scores. Outputs that are vague or general are scored near $0$ or slightly negative, whereas irrelevant or nonsensical outputs receive scores close to $-10$. Several detail cases are listed in \Cref{tab:rm-examples}.

\begin{table}[t]
\centering
\caption{Examples of summary alignment reward.}
\label{tab:rm-examples}
\setlength{\tabcolsep}{3pt}
\small
\begin{tabular}{@{}c p{0.85\linewidth} r@{}}
\toprule
\multicolumn{2}{@{}p{0.9\linewidth}@{}}{\textbf{Reference:} \textit{Make the image brighter and vivid, and add sharpness to it.}}  & \multirow{2}{*}{\textbf{$R_A$}} \\
\midrule
1) & The user prefers a brighter and more colorful image, and also makes the image sharper. & \textbf{10} \\
2) & The user prefers a much brighter and less colorful image. & \textbf{-5} \\
3) & The user prefers an image with high contrast and sharpness. & \textbf{3} \\
4) & The user wants to make the image nicer. & \textbf{0} \\
5) & Today is a sunny day. (Irrelevant answer, or in a different language, or kept repeating the same sentence, or other nonsense) & \textbf{-10} \\
\bottomrule
\end{tabular}
\end{table}

\subsection{Stage 3: Policy Generalization}
\label{sec:stage3}

\paragraph{Objective.} The filtered training samples extracted from GIER exhibit limited coverage of the available toolset---only about half of the 16 supported tools in our built-in editor are actively used---and contain too few “same image, different intent” pairs to adequately supervise refinement tasks. To address these deficiencies, we synthesize additional data by sampling subsets of tools and parameter values from the editor, generating corresponding template-based instructions, and paraphrasing them into natural, amateur-style commands using \texttt{GPT-4.1}. This augmentation strategy: (i) ensures full activation of all 16 tools as well as their interactions; (ii) exposes the model to a wide spectrum of instruction styles ranging from casual amateur queries to precise expert directives; and (iii) facilitates straightforward construction of similar-instruction pairs for alignment RM training and refinement task generation.

\vspace{-0.5cm}
\paragraph{Synthesis Pipeline.} We follow the steps below to synthesize all data for Stage 3 and alignment RM training. \textbf{(1) Tool selection.} We sample a subset of 16 editor-supported tools and assign parameter magnitudes from qualitatively defined bins (e.g., \textit{slightly increase}, \textit{moderately decrease}, \textit{strongly increase}, etc.). Each qualitative magnitude is also mapped to a precise numerical value. These tool–magnitude combinations define a latent expert program $T^\star$. \textbf{(2) Instruction generation and paraphrasing.} From $T^\star$, we first generate an expert-style instruction $Q_{\text{expert}}^\star$ using a predefined template (e.g., \textit{``Increase exposure significantly, add slight sharpness, and warm the image."}). We then paraphrase it with \texttt{GPT-4.1} into a natural, amateur-style request $Q_{\text{amateur}}^\star$ that emphasizes visual effects rather than explicit parameters (e.g., \textit{``Make it much brighter, slightly sharper, and warmer—keep the colors pleasant."}). This dual representation of user intent enhances linguistic diversity while preserving executable semantics. \textbf{(3) Image–label construction.} Each image is randomly selected from either the GIER dataset~\cite{gier} or the FiveK dataset~\cite{fivek}. Given an original image $I_{\text{ori}}$, we apply $T^\star$ using the editor to obtain $I^\star = \mathcal{E}(I_{\text{ori}}, T^\star)$. \textbf{(4) Training data synthesis.} Each synthetic sample consists of $(I_{\text{ori}}, I^\star, Q_{\text{expert}}^\star, Q_{\text {amateur}}^\star, T^\star)$. These synthesized pairs are then used to construct Image-Summary-Synthesis and Image-Edit-Synthesis datasets following the same procedure described in Stage 1. \textbf{(5) Similar instruction generation.} Since all instructions originate from an “expert program” $T^\star$, we can easily create variations by adding or removing tools or adjusting magnitudes to form a new expert program $T^{\star\prime}$. A relevance score is assigned to each $(T^\star, T^{\star\prime})$ pair as the target alignment reward for training the RM used in Stage 2.

\paragraph{Image-Refine Task Formulation}  
This task is designed to enable the model to interpret a user's updated instruction and refine the selected set of tools accordingly. We sample two related programs, $(T^{(0)}, Q^{(0)})$ and $(T^{(1)}, Q^{(1)})$, which differ in one or two dimensions (e.g., $T^{(0)}$ applies \textit{strong} exposure, whereas $T^{(1)}$ reduces it to \textit{moderate}). We then render:
\vspace{-0.2cm}
\begin{equation}
I^{(0)} = \mathcal{E}(I_{\text{ori}}, T^{(0)}), \qquad
I^{(1)} = \mathcal{E}(I_{\text{ori}}, T^{(1)}),
\end{equation}
Next, we prompt \texttt{GPT-4.1} to generate a concise refinement instruction $\Delta Q$—expressed in natural language  (e.g., ``The brightness is a bit too strong—dial it down slightly.")—that would transform $I^{(0)}$ toward $I^{(1)}$, given $(T^{(0)}, T^{(1)}, Q^{(0)}, Q^{(1)})$. The learning objective is then defined as predicting the refined tool parameters from this refinement context:
\vspace{-0.2cm}
\begin{equation}
\mathcal{M}\big(I_{\text{ori}}, Q^{(0)}, T^{(0)}, I^{(0)}, \Delta Q\big) \rightarrow T^{(1)}.
\end{equation}
\vspace{-0.5cm}

\noindent Here, $Q^{(0)}$ denotes the initial user instruction that produced $T^{(0)}$. This formulation trains the policy to perform incremental edits while faithfully incorporating user feedback.

\begin{figure}[h]
    \centering
    \includegraphics[width=\linewidth]{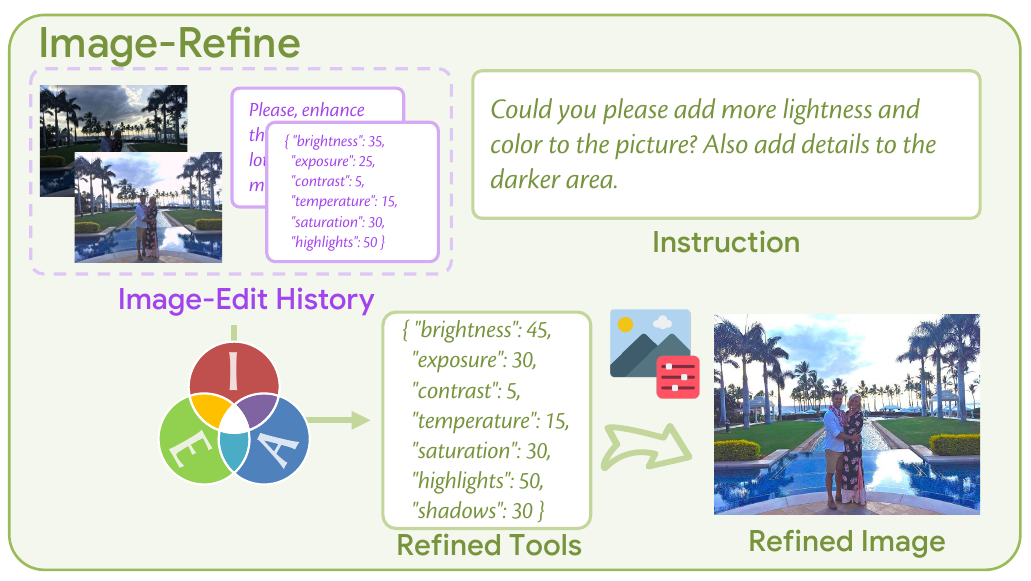}
    \caption{Example of Image-Refine Task.}
    \label{fig:image-refine}
\end{figure}
\vspace{-0.5cm}

\section{Quantitative Experiments}

\paragraph{Setup.} We fine-tune \texttt{Qwen2.5-VL-7B} as the unified vision–language policy. We use a batch size of 128, learning rate $1{\times}10^{-5}$, and train for 5 epochs on $\sim$29k SFT examples, totaling $\sim$1.1k optimization steps on 4 Nvidia-A800 GPUs. Finally, we obtain the finetuned model as \texttt{IEA-Stage-1}.
 
We adopt the \texttt{verl}~\cite{sheng2024hybridflow} training framework to perform GRPO training on \texttt{IEA-Stage-1}. We use a batch size of 16, and train for 1 epoch on $\sim$29k RL examples on 8 Nvidia-A800 GPUs, while retaining the exact SFT output formats. We set weight $\alpha$ to 0.7 and rollout number to $K=5$. Finally, we obtain the actor model as \texttt{IEA-Stage-2}. Please refer to the supplementary material for details of the alignment RM training.

We interleave real and synthetic supervision to form $\sim$132k synthetic SFT items covering three tasks. In addition, we re-sample $\sim$70k original SFT items from Stage~1, yielding a mixed corpus of $\sim$202k examples. Please refer to the supplementary material for detailed data composition.

We fine-tune the \texttt{IEA-Stage-2} policy with a batch size of 64 and learning rate $1{\times}10^{-5}$ for 2 epochs (approximately 5k steps) on 8 Nvidia-A800 GPUs. Output formats remain identical to Stage~1/2 to minimize distributional shift: JSON with fixed keys for \textit{Image-Edit}/\textit{Image-Refine}, and a single English sentence for \textit{Image-Summary}. The resulting model is denoted as \texttt{IEA-Stage-3}.

To quantitatively evaluate the effectiveness of \agent{}, we conduct experiments on two tasks: \textit{Image-Edit} and \textit{Image-Summary}. For fair comparison, we select 2,568 test samples for \textit{Image-Edit} and 241 test samples for \textit{Image-Summary}. We use frontier large language models (\texttt{GPT-4.1}~\cite{gpt4.1}, \texttt{Gemini-2.5-Pro}~\cite{gemini2.5}) and the backbone model (\texttt{Qwen2.5-VL-7B}~\cite{Qwen2.5-VL}) as our baselines.

\vspace{-0.3cm}
\paragraph{Metrics.} We report rewards and scores pre-defined in \Cref{sec:stage2}. For \textit{Image-Edit}, we report pixel-level distance $L=\mathcal{L}(I_{\text{edit}},I_{\text{ref}})$ (averaging L1 and L2), likeness improvement reward $R_L$, and usefulness reward $R_U$. For \textit{Image-Summary}, we compute Rouge-L~\cite{lin-2004-rouge} against the ground-truth textual preference and alignment reward $R_A$ from our trained reward model.

\begin{table}[t]
\centering
\caption{Experiment results in Image-Edit and Image-Summary tasks.} 
\label{tab:experiments}
\setlength{\tabcolsep}{2pt}
\renewcommand{\arraystretch}{1.15}
\small
\begin{tabular}{lccccc}
\toprule
\multirow{2.5}{*}{\textbf{Model}} & \multicolumn{3}{c}{\textbf{Image-Edit}} & \multicolumn{2}{c}{\textbf{Image-Summary}} \\
\cmidrule(lr){2-4} \cmidrule(lr){5-6}
 & \textbf{L } (↓)& \textbf{$R_L$ } (↑)& \textbf{$R_U$ }(↑) & \textbf{Rouge-L } (↑)& \textbf{$R_A$ }(↑)  \\  
\midrule
GPT-4.1              & 0.150 & -0.296 & -0.408 & 0.097 & -0.441 \\
Gemini-2.5-Pro        & 0.168 & -0.415 & -0.301 & 0.077 & 2.399 \\
QwenVL-2.5-7B         & 0.158 & -0.284 & -0.412 & 0.091 & 3.671 \\
\midrule
IEA-Stage-1           & 0.134 & -0.219 & -0.104 & 0.222 & 3.852 \\
IEA-Stage-2           & 0.111 & 0.0271 & 0.332 & 0.196 & 4.859 \\
IEA-Stage-3           & 0.103 & 0.149 & 0.402 & 0.258 & 7.387 \\
\bottomrule
\end{tabular}
\end{table}

As shown in Table~\ref{tab:experiments}, our staged pipeline consistently improves performance across both tasks. \textit{IEA-Stage-1} significantly reduces $L$ and increases Rouge-L compared to the base \textit{Qwen2.5-VL-7B}, indicating that semi-supervised distillation effectively transfers tool-usage knowledge of \textit{what to do} and \textit{what has been done}. With reinforcement learning, \textit{IEA-Stage-2} achieves positive $R_L$ and $R_U$, showing better alignment with reference edits and more efficient tool use. Finally, \textit{IEA-Stage-3}, which leverages large-scale synthetic data, further reduces pixel distance and achieves the highest alignment reward $R_A$, nearly doubling that of Stage-1. Compared with frontier generative models, \agent{} not only achieves lower edit distance but also produces more instruction-consistent summaries, validating the advantage of parametric editing and preference alignment.
\section{Qualitative Studies}

To provide a concrete illustration of \agent{}'s capabilities with other baselines, we present a comparative user study on 50 examples from the GIER dataset. We compared \agent{} against both generative methods and tool-calling methods of image editing:
\begin{itemize}
    \item \textbf{Generative Methods:} These models directly synthesize pixels using diffusion models in an end-to-end manner. This category includes \texttt{GPT-Image-1}~\cite{gpt-image}, and \texttt{Qwen-Image-Edit}~\cite{Qwen2.5-VL}.
    \item \textbf{Tool-Calling Methods:} These models generate parameters for predefined editing tools, and produce edited images afterward. We select \texttt{Gemini-2.5-Pro}~\cite{gemini2.5}, \texttt{GPT-4.1}~\cite{gpt4.1}, and the base \texttt{Qwen2.5-VL-7B}~\cite{Qwen2.5-VL} model for direct comparison with \agent{}. For these models, we use the same pipeline as \agent{}, but also include format instructions and the available tool list in the prompt. We also select \texttt{JarvisArt-Preview} (trained from \texttt{Qwen2.5-VL-7B}) as a baseline, which accepts a user instruction and original image, and generates a format template file that could be used in Adobe Lightroom. We then manually process each image with the corresponding template to get the final image. 
    \item We also include reference images annotated by human experts from the GIER dataset.
\end{itemize}

\subsection{User Study}\label{sec:user_study}

To evaluate the performance of our proposed \agent{} system, we conducted a user study comparing it against several baselines. The study was designed to assess both instruction-following capability and perceptual quality of the edited images.

\paragraph{Evaluation Tasks.} Participants were presented with the original image and a text instruction, both in English and Simplified Chinese. They were then asked to evaluate and rank the edited outputs of all methods (including the human expert reference) in two tasks:

\textbf{Task A: Instruction Following Assessment.}\
They were asked to rank the images based on how accurately each result followed the given instruction. Completely disregarding the instructions, or excessively following them (such as "slightly increasing brightness" but actually increasing it so much that the image is overexposed), will result in a lower ranking.

\textbf{Task B: Image Quality Assessment.}\
They were asked to rank the same set of images based on overall perceptual quality. Participants considered factors such as natural appearance, visual authenticity, absence of artifacts, and overall aesthetic appeal. 

\paragraph{Annotation Detail.} We recruited 56 participants from the university who are able to understand instructions in both Simplified Chinese and English, have basic image editing experience, and have no visual impairments. Their compensation meets local standards, and personal privacy is ensured. Please refer to supplementary material for more.

\paragraph{Results.}
\begin{figure}[t]
    \centering
    \begin{subfigure}[h]{\linewidth}
        \centering
        \includegraphics[width=\linewidth]{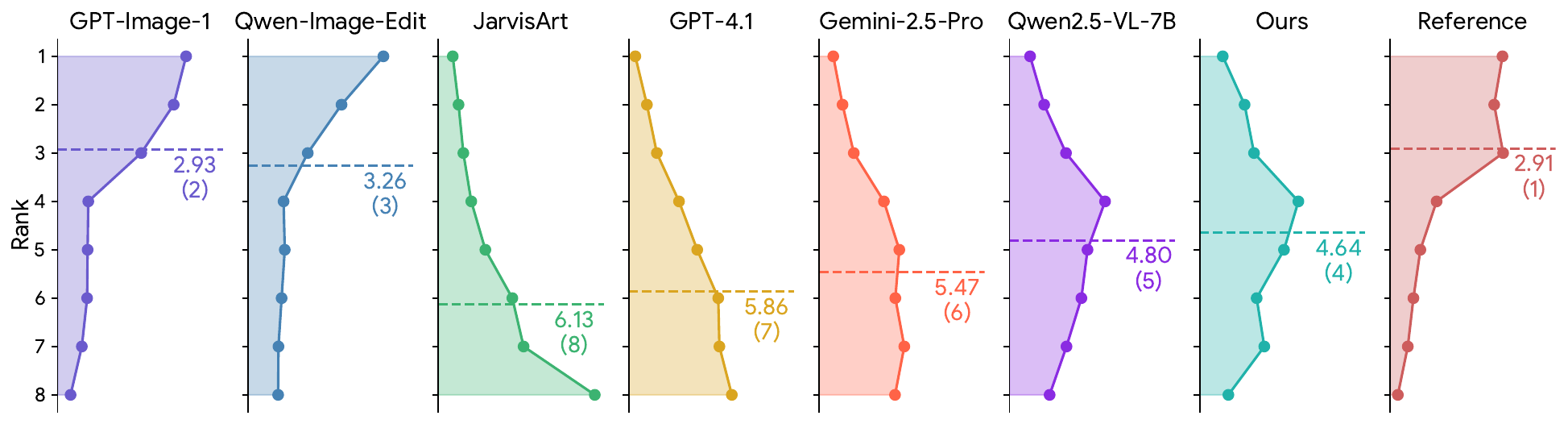}
        \caption{Task A: Instruction Following Assessment.}
        \label{fig:A_rank_distribution}
    \end{subfigure}
    

    \begin{subfigure}[h]{\linewidth}
        \centering
        \includegraphics[width=\linewidth]{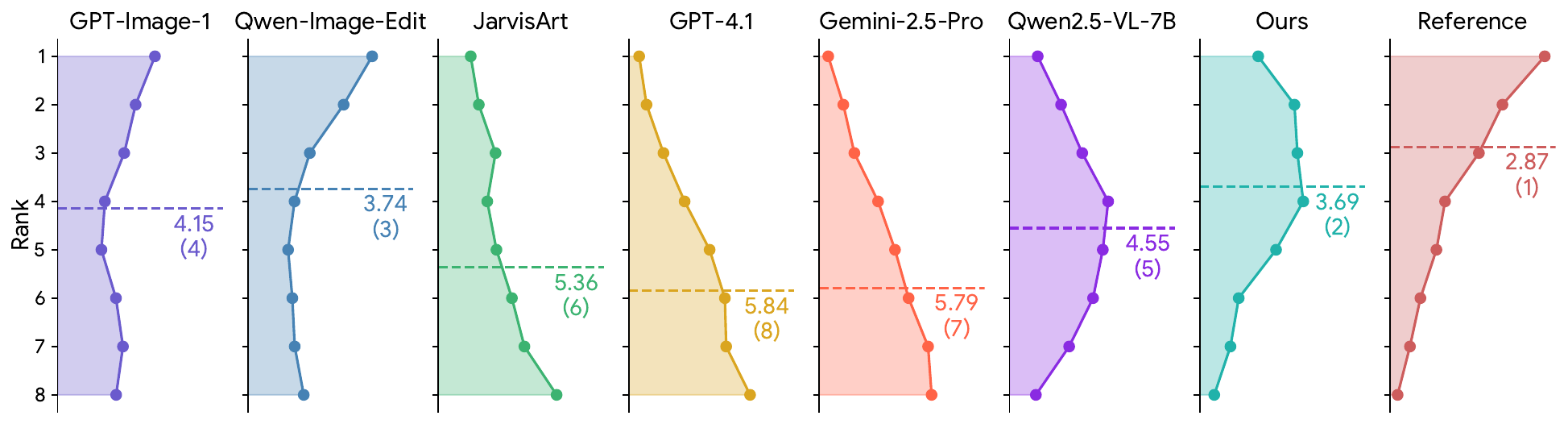}
        \caption{Task B: Image Quality Assessment.}
        \label{fig:B_rank_distribution}
    \end{subfigure}

    \caption{Comparison of rank distributions for two tasks. The dashed lines indicate the average ranks of each model, and the number in brackets is the rank of the average ranks.}
    \label{fig:rank_distribution}
\end{figure}

\begin{table}[t]
\centering
\caption{Results on the 50 user study samples. The best results are \textbf{bolded}, and the second-best results are \underline{underlined}. Rank(A): The average rank of instruction following assessment, Rank(B): The average rank of image quality assessment.}
\label{tab:all-results}
\setlength{\tabcolsep}{8pt}
\renewcommand{\arraystretch}{1}
\small
\begin{tabular}{lcccc}
\toprule
\textbf{Model} & \makecell{\textbf{Rank}\\\textbf{(A$\downarrow$)}}& \makecell{\textbf{Rank}\\\textbf{(B$\downarrow$)}}& \textbf{L} ($\downarrow$) & \textbf{$R_L$} ($\uparrow$) \\
\midrule
Reference & 2.91 & 2.87 & 0.000 & 1.000 \\
\midrule
GPT-Image-1 & \textbf{2.93} & 4.15 & 0.218 & -0.340 \\
Qwen-Image-Edit & \underline{3.26} & \underline{3.74} & 0.184 & \underline{-0.132} \\
JarvisArt & 6.13 & 5.36 & 0.225 & -0.328 \\
GPT-4.1 & 5.86 & 5.84 & 0.209 & -0.306 \\
Gemini-2.5-Pro & 5.47 & 5.79 & 0.225 & -0.420 \\
Qwen2.5-VL-7B & 4.80 & 4.55 & \underline{0.200} & -0.221 \\
\midrule
Ours & 4.64 & \textbf{3.69} & \textbf{0.128} & \textbf{0.138} \\
\bottomrule
\end{tabular}
\end{table}

For the instruction-following task, \agent{} outperforms other tool-calling baselines. With SFT and GRPO, it exploits the constrained action space of image-editing tools to approximate human-expert edits as closely as possible. By contrast, generative models can regenerate pixels directly according to the instruction; in severely under-/over-exposed scenes or monochrome photos, they can plausibly complete details that are unrecoverable by tools, which sometimes yields higher ranks.

For the image-quality task, \agent{} produces stable and natural results by selecting appropriate tools and calibrated parameter values—satisfying the instruction while avoiding over-editing, erroneous tool invocations, and other failure modes that cause visual anomalies. Generative models, however, more often introduce artifacts, implausible details, or stylistic drift away from photorealism, leading to inferior overall perceptual quality compared with tool-based, pixel-level retouching.

\subsection{Case Study}
\label{sec:case_study}

\begin{figure}[h]
\centering
\includegraphics[width=\linewidth,trim=5 0 25 0,clip]{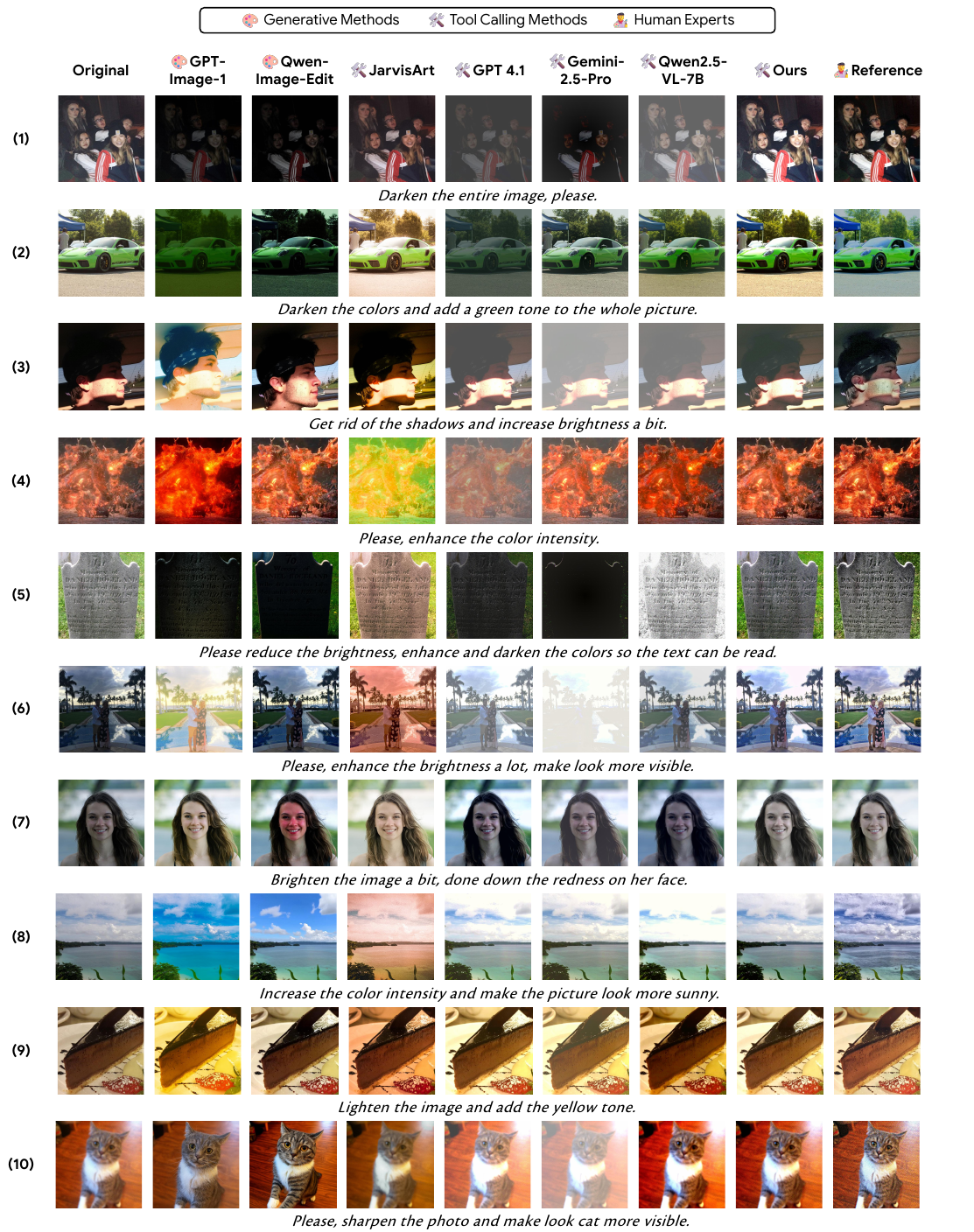} 
\caption{Comparison of image editing results. Images are cropped and resized to 512-by-512 pixels for clear presentation. Readers can zoom in to examine the details.}
\label{fig:comparison}
\end{figure}

\Cref{fig:comparison} presents the editing results generated by various baseline methods alongside those produced by \agent{}, using the same original images and corresponding editing instructions.

\textbf{Example 1: ``Darken the entire image, please''}
The original image is a slightly overexposed nighttime group photo. Generative models often struggle with such precise, non-generative tasks, potentially altering content or introducing artifacts instead of performing a global adjustment. Other parametric models frequently apply excessive adjustments, further obscuring the faces of the people in the image. This suggests a fundamental difficulty in precisely mapping natural language to appropriate parameter values. In contrast, our model applies a more measured darkening, successfully improving visibility and preserving details of the subjects.

\textbf{Example 4: ``Please, enhance the color intensity''}
This instruction requires enhancing color without compromising image fidelity. Generative models, relying on pixel synthesis, frequently suffer from significant loss of fine details, resulting in edited images that appear blurry or lack texture—a common trade-off for their generality. Other tool-calling models often fail to grasp the nuanced meaning of "enhance," resulting in either insufficient change or an oversaturated, unrealistic look. Our model effectively boosts the color vibrancy, making the image more vivid and lively while maintaining its structural integrity and realism. 

\textbf{Example 7: ``Brighten the image a bit, tone down the redness on her face''}
This complex instruction requires a nuanced understanding of both global and local attributes. Our model and the GIER reference both successfully achieve a subtle brightening that enhances overall clarity without disrupting the original color temperature of the scene. Other parametric models either produce negligible changes or apply global adjustments that are too drastic, altering the fundamental tone of the entire image and failing to address the localized issue.

The comparison has validated that \agent{} not only follows instructions more accurately, but also produces more natural and aesthetically pleasing results than competing methods.

\section{Conclusion}
In this work, we introduced \agent{}, an amateur-friendly conversational image editing agent that operates parameterized tools instead of synthesizing pixels. Built on a simulated editor and a three-stage multitask alignment pipeline—SFT on distilled expert edits, GRPO with likeness, usefulness, and summary rewards, and large-scale synthetic supervision—\agent{} learns to perform Image-Edit, Image-Summary, and Image-Refine in a unified framework. Experiments on GIER and FiveK show that \agent{} consistently lowers pixel distance to expert references, improves preference summarization scores, and, in user studies, ranks highest among tool-based methods for instruction following while surpassing generative methods in overall perceptual quality. Despite being trained with 16 global tools in a simplified environment, \agent{} demonstrates that tool-centric VLMs can deliver interpretable, preference-aligned photo retouching and offers a foundation for future work toward richer tools, more realistic backends, and human-in-the-loop refinement.

\section*{Acknowledgments}
This work was supported by the China NSFC Projects (62576212, 92370206, U23B2057, 62120106006).

{
    \small
    \bibliographystyle{ieeenat_fullname}
    \bibliography{main,cite}

\begin{thebibliography}{81}
\providecommand{\natexlab}[1]{#1}
\providecommand{\url}[1]{\texttt{#1}}
\expandafter\ifx\csname urlstyle\endcsname\relax
  \providecommand{\doi}[1]{doi: #1}\else
  \providecommand{\doi}{doi: \begingroup \urlstyle{rm}\Url}\fi

\bibitem[Ahn et~al.(2022)Ahn, Brohan, Brown, Chebotar, Cortes, David, Finn, Fu, Gopalakrishnan, Hausman, Herzog, Ho, Hsu, Ibarz, Ichter, Irpan, Jang, Ruano, Jeffrey, Jesmonth, Joshi, Julian, Kalashnikov, Kuang, Lee, Levine, Lu, Luu, Parada, Pastor, Quiambao, Rao, Rettinghouse, Reyes, Sermanet, Sievers, Tan, Toshev, Vanhoucke, Xia, Xiao, Xu, Xu, Yan, and Zeng]{saycan2022arxiv}
Michael Ahn, Anthony Brohan, Noah Brown, Yevgen Chebotar, Omar Cortes, Byron David, Chelsea Finn, Chuyuan Fu, Keerthana Gopalakrishnan, Karol Hausman, Alex Herzog, Daniel Ho, Jasmine Hsu, Julian Ibarz, Brian Ichter, Alex Irpan, Eric Jang, Rosario~Jauregui Ruano, Kyle Jeffrey, Sally Jesmonth, Nikhil Joshi, Ryan Julian, Dmitry Kalashnikov, Yuheng Kuang, Kuang-Huei Lee, Sergey Levine, Yao Lu, Linda Luu, Carolina Parada, Peter Pastor, Jornell Quiambao, Kanishka Rao, Jarek Rettinghouse, Diego Reyes, Pierre Sermanet, Nicolas Sievers, Clayton Tan, Alexander Toshev, Vincent Vanhoucke, Fei Xia, Ted Xiao, Peng Xu, Sichun Xu, Mengyuan Yan, and Andy Zeng.
\newblock Do as i can and not as i say: Grounding language in robotic affordances.
\newblock In \emph{arXiv preprint arXiv:2204.01691}, 2022.

\bibitem[Avrahami et~al.(2022)Avrahami, Lischinski, and Fried]{Avrahami_2022_CVPR}
Omri Avrahami, Dani Lischinski, and Ohad Fried.
\newblock Blended diffusion for text-driven editing of natural images.
\newblock In \emph{Proceedings of the IEEE/CVF Conference on Computer Vision and Pattern Recognition (CVPR)}, pages 18208--18218, 2022.

\bibitem[Bai et~al.(2025)Bai, Chen, Liu, Wang, Ge, Song, Dang, Wang, Wang, Tang, Zhong, Zhu, Yang, Li, Wan, Wang, Ding, Fu, Xu, Ye, Zhang, Xie, Cheng, Zhang, Yang, Xu, and Lin]{Qwen2.5-VL}
Shuai Bai, Keqin Chen, Xuejing Liu, Jialin Wang, Wenbin Ge, Sibo Song, Kai Dang, Peng Wang, Shijie Wang, Jun Tang, Humen Zhong, Yuanzhi Zhu, Mingkun Yang, Zhaohai Li, Jianqiang Wan, Pengfei Wang, Wei Ding, Zheren Fu, Yiheng Xu, Jiabo Ye, Xi Zhang, Tianbao Xie, Zesen Cheng, Hang Zhang, Zhibo Yang, Haiyang Xu, and Junyang Lin.
\newblock Qwen2.5-vl technical report.
\newblock \emph{arXiv preprint arXiv:2502.13923}, 2025.

\bibitem[Brooks et~al.(2023)Brooks, Holynski, and Efros]{brooks2023instructpix2pix}
Tim Brooks, Aleksander Holynski, and Alexei~A Efros.
\newblock Instructpix2pix: Learning to follow image editing instructions.
\newblock In \emph{Proceedings of the IEEE/CVF conference on computer vision and pattern recognition}, pages 18392--18402, 2023.

\bibitem[Bychkovsky et~al.(2011)Bychkovsky, Paris, Chan, and Durand]{fivek}
Vladimir Bychkovsky, Sylvain Paris, Eric Chan, and Fr{\'e}do Durand.
\newblock Learning photographic global tonal adjustment with a database of input / output image pairs.
\newblock In \emph{The Twenty-Fourth IEEE Conference on Computer Vision and Pattern Recognition}, 2011.

\bibitem[Chen et~al.(2025)Chen, Tao, Wang, Wang, Zhu, and Gu]{chen2025photoartagent}
Haoyu Chen, Keda Tao, Yizao Wang, Xinlei Wang, Lei Zhu, and Jinjin Gu.
\newblock Photoartagent: Intelligent photo retouching with language model-based artist agents.
\newblock \emph{arXiv preprint arXiv:2505.23130}, 2025.

\bibitem[Comanici et~al.(2025)Comanici, Bieber, Schaekermann, Pasupat, Sachdeva, Dhillon, Blistein, Ram, Zhang, Rosen, et~al.]{gemini2.5}
Gheorghe Comanici, Eric Bieber, Mike Schaekermann, Ice Pasupat, Noveen Sachdeva, Inderjit Dhillon, Marcel Blistein, Ori Ram, Dan Zhang, Evan Rosen, et~al.
\newblock Gemini 2.5: Pushing the frontier with advanced reasoning, multimodality, long context, and next generation agentic capabilities.
\newblock \emph{arXiv preprint arXiv:2507.06261}, 2025.

\bibitem[Conde et~al.(2025)Conde, Lu, and Timofte]{Conde_2025_ICCV}
Marcos~V. Conde, Zihao Lu, and Radu Timofte.
\newblock Pixtalk: Controlling photorealistic image processing and editing with language.
\newblock In \emph{Proceedings of the IEEE/CVF International Conference on Computer Vision (ICCV)}, pages 19269--19279, 2025.

\bibitem[Deutch et~al.(2024)Deutch, Gal, Garibi, Patashnik, and Cohen-Or]{deutch2024turboedittextbasedimageediting}
Gilad Deutch, Rinon Gal, Daniel Garibi, Or Patashnik, and Daniel Cohen-Or.
\newblock Turboedit: Text-based image editing using few-step diffusion models, 2024.

\bibitem[Fan et~al.(2024)Fan, Ding, Ning, Wang, Li, Yin, Chua, and Li]{10.1145/3637528.3671470}
Wenqi Fan, Yujuan Ding, Liangbo Ning, Shijie Wang, Hengyun Li, Dawei Yin, Tat-Seng Chua, and Qing Li.
\newblock A survey on rag meeting llms: Towards retrieval-augmented large language models.
\newblock In \emph{Proceedings of the 30th ACM SIGKDD Conference on Knowledge Discovery and Data Mining}, page 6491–6501, New York, NY, USA, 2024. Association for Computing Machinery.

\bibitem[Fu et~al.(2024)Fu, Hu, Du, Wang, Yang, and Gan]{fu2024guiding}
Tsu-Jui Fu, Wenze Hu, Xianzhi Du, William~Yang Wang, Yinfei Yang, and Zhe Gan.
\newblock Guiding instruction-based image editing via multimodal large language models.
\newblock In \emph{The Twelfth International Conference on Learning Representations}, 2024.

\bibitem[Gao et~al.(2025)Gao, Dwivedi-Yu, Yu, Tan, Pasunuru, Golovneva, Sinha, Celikyilmaz, Bosselut, and Wang]{gao-etal-2025-efficient}
Silin Gao, Jane Dwivedi-Yu, Ping Yu, Xiaoqing~Ellen Tan, Ramakanth Pasunuru, Olga Golovneva, Koustuv Sinha, Asli Celikyilmaz, Antoine Bosselut, and Tianlu Wang.
\newblock Efficient tool use with chain-of-abstraction reasoning.
\newblock In \emph{Proceedings of the 31st International Conference on Computational Linguistics}, pages 2727--2743, Abu Dhabi, UAE, 2025. Association for Computational Linguistics.

\bibitem[Hao et~al.(2023)Hao, Liu, Wang, and Hu]{hao2023toolkengpt}
Shibo Hao, Tianyang Liu, Zhen Wang, and Zhiting Hu.
\newblock Toolken{GPT}: Augmenting frozen language models with massive tools via tool embeddings.
\newblock In \emph{Thirty-seventh Conference on Neural Information Processing Systems}, 2023.

\bibitem[Hertz et~al.(2022)Hertz, Mokady, Tenenbaum, Aberman, Pritch, and Cohen-Or]{hertz2022prompt}
Amir Hertz, Ron Mokady, Jay Tenenbaum, Kfir Aberman, Yael Pritch, and Daniel Cohen-Or.
\newblock Prompt-to-prompt image editing with cross attention control.
\newblock \emph{arXiv preprint arXiv:2208.01626}, 2022.

\bibitem[Hu et~al.(2018)Hu, He, Xu, Wang, and Lin]{hu2018exposure}
Yuanming Hu, Hao He, Chenxi Xu, Baoyuan Wang, and Stephen Lin.
\newblock Exposure: A white-box photo post-processing framework.
\newblock \emph{ACM Transactions on Graphics (TOG)}, 37\penalty0 (2):\penalty0 1--17, 2018.

\bibitem[Huang et~al.(2025)Huang, Chan, Liu, He, Jiang, Song, and Song]{huang2025patchdpo}
Qihan Huang, Long Chan, Jinlong Liu, Wanggui He, Hao Jiang, Mingli Song, and Jie Song.
\newblock Patch{DPO}: Patch-level dpo for finetuning-free personalized image generation.
\newblock In \emph{Proceedings of the Computer Vision and Pattern Recognition Conference}, pages 18369--18378, 2025.

\bibitem[Huang et~al.(2024)Huang, Xie, Wang, Yuan, Cun, Ge, Zhou, Dong, Huang, Zhang, and Shan]{Huang_2024_CVPR}
Yuzhou Huang, Liangbin Xie, Xintao Wang, Ziyang Yuan, Xiaodong Cun, Yixiao Ge, Jiantao Zhou, Chao Dong, Rui Huang, Ruimao Zhang, and Ying Shan.
\newblock Smartedit: Exploring complex instruction-based image editing with multimodal large language models.
\newblock In \emph{Proceedings of the IEEE/CVF Conference on Computer Vision and Pattern Recognition (CVPR)}, pages 8362--8371, 2024.

\bibitem[Ji et~al.(2023)Ji, Lee, Frieske, Yu, Su, Xu, Ishii, Bang, Madotto, and Fung]{10.1145/3571730}
Ziwei Ji, Nayeon Lee, Rita Frieske, Tiezheng Yu, Dan Su, Yan Xu, Etsuko Ishii, Ye~Jin Bang, Andrea Madotto, and Pascale Fung.
\newblock Survey of hallucination in natural language generation.
\newblock \emph{ACM Comput. Surv.}, 55\penalty0 (12), 2023.

\bibitem[Kawar et~al.(2023)Kawar, Zada, Lang, Tov, Chang, Dekel, Mosseri, and Irani]{DBLP:conf/cvpr/KawarZLTCDMI23}
Bahjat Kawar, Shiran Zada, Oran Lang, Omer Tov, Huiwen Chang, Tali Dekel, Inbar Mosseri, and Michal Irani.
\newblock Imagic: Text-based real image editing with diffusion models.
\newblock In \emph{CVPR}, pages 6007--6017, 2023.

\bibitem[Ke et~al.(2023)Ke, Liu, Zhu, Zhao, and Lau]{Ke_2023_CVPR}
Zhanghan Ke, Yuhao Liu, Lei Zhu, Nanxuan Zhao, and Rynson~W.H. Lau.
\newblock Neural preset for color style transfer.
\newblock In \emph{Proceedings of the IEEE/CVF Conference on Computer Vision and Pattern Recognition (CVPR)}, pages 14173--14182, 2023.

\bibitem[Kingma and Welling(2013)]{kingma2013auto}
Diederik~P Kingma and Max Welling.
\newblock Auto-encoding variational bayes.
\newblock \emph{arXiv preprint arXiv:1312.6114}, 2013.

\bibitem[Krojer et~al.(2024)Krojer, Vattikonda, Lara, Jampani, Portelance, Pal, and Reddy]{krojer2024aurora}
Benno Krojer, Dheeraj Vattikonda, Luis Lara, Varun Jampani, Eva Portelance, Christopher Pal, and Siva Reddy.
\newblock {Learning Action and Reasoning-Centric Image Editing from Videos and Simulations}.
\newblock In \emph{NeurIPS}, 2024.
\newblock Spotlight Paper.

\bibitem[Lan et~al.(2024)Lan, Jin, Zhu, Chen, Zhang, Zhu, and Wu]{lan2024depressiondiagnosisdialoguesimulation}
Kunyao Lan, Bingrui Jin, Zichen Zhu, Siyuan Chen, Shu Zhang, Kenny~Q. Zhu, and Mengyue Wu.
\newblock Depression diagnosis dialogue simulation: Self-improving psychiatrist with tertiary memory, 2024.

\bibitem[Lee et~al.(2024)Lee, Kang, Ok, and Cho]{Lee_2024_CVPR}
Hyeongmin Lee, Kyoungkook Kang, Jungseul Ok, and Sunghyun Cho.
\newblock Cliptone: Unsupervised learning for text-based image tone adjustment.
\newblock In \emph{Proceedings of the IEEE/CVF Conference on Computer Vision and Pattern Recognition (CVPR)}, pages 2942--2951, 2024.

\bibitem[Lei et~al.(2023)Lei, Ding, Cao, Zan, Yates, and Tao]{lei-etal-2023-unsupervised}
Yibin Lei, Liang Ding, Yu Cao, Changtong Zan, Andrew Yates, and Dacheng Tao.
\newblock Unsupervised dense retrieval with relevance-aware contrastive pre-training.
\newblock In \emph{Findings of the Association for Computational Linguistics: ACL 2023}, pages 10932--10940, Toronto, Canada, 2023. Association for Computational Linguistics.

\bibitem[Li et~al.(2023)Li, Zhao, Yu, Song, Li, Yu, Li, Huang, and Li]{DBLP:conf/emnlp/LiZ000YLHL23}
Minghao Li, Yingxiu Zhao, Bowen Yu, Feifan Song, Hangyu Li, Haiyang Yu, Zhoujun Li, Fei Huang, and Yongbin Li.
\newblock Api-bank: A comprehensive benchmark for tool-augmented llms.
\newblock In \emph{EMNLP}, pages 3102--3116, 2023.

\bibitem[Lin(2004)]{lin-2004-rouge}
Chin-Yew Lin.
\newblock {ROUGE}: A package for automatic evaluation of summaries.
\newblock In \emph{Text Summarization Branches Out}, pages 74--81, Barcelona, Spain, 2004. Association for Computational Linguistics.

\bibitem[Lin et~al.(2025)Lin, Lin, Lin, Bai, Pan, Li, Chen, Wang, Ding, Li, et~al.]{lin2025jarvisart}
Yunlong Lin, Zixu Lin, Kunjie Lin, Jinbin Bai, Panwang Pan, Chenxin Li, Haoyu Chen, Zhongdao Wang, Xinghao Ding, Wenbo Li, et~al.
\newblock Jarvisart: Liberating human artistic creativity via an intelligent photo retouching agent.
\newblock \emph{arXiv preprint arXiv:2506.17612}, 2025.

\bibitem[Liu et~al.(2024{\natexlab{a}})Liu, Peng, Yi, Xie, Xiang, Liu, and Xu]{DBLP:journals/corr/abs-2403-00839}
Xukun Liu, Zhiyuan Peng, Xiaoyuan Yi, Xing Xie, Lirong Xiang, Yuchen Liu, and Dongkuan Xu.
\newblock Toolnet: Connecting large language models with massive tools via tool graph.
\newblock \emph{CoRR}, abs/2403.00839, 2024{\natexlab{a}}.

\bibitem[Liu et~al.(2024{\natexlab{b}})Liu, Yu, Zhang, Xu, Lei, Lai, Gu, Ding, Men, Yang, Zhang, Deng, Zeng, Du, Zhang, Shen, Zhang, Su, Sun, Huang, Dong, and Tang]{liu2024agentbench}
Xiao Liu, Hao Yu, Hanchen Zhang, Yifan Xu, Xuanyu Lei, Hanyu Lai, Yu Gu, Hangliang Ding, Kaiwen Men, Kejuan Yang, Shudan Zhang, Xiang Deng, Aohan Zeng, Zhengxiao Du, Chenhui Zhang, Sheng Shen, Tianjun Zhang, Yu Su, Huan Sun, Minlie Huang, Yuxiao Dong, and Jie Tang.
\newblock Agentbench: Evaluating {LLM}s as agents.
\newblock In \emph{The Twelfth International Conference on Learning Representations}, 2024{\natexlab{b}}.

\bibitem[Liu et~al.(2024{\natexlab{c}})Liu, Lai, Gao, Cui, Li, Zhu, Lu, Chen, Qiao, Dai, and Wang]{10.1007/978-3-031-73254-6_6}
Zhaoyang Liu, Zeqiang Lai, Zhangwei Gao, Erfei Cui, Ziheng Li, Xizhou Zhu, Lewei Lu, Qifeng Chen, Yu Qiao, Jifeng Dai, and Wenhai Wang.
\newblock Controlllm: Augment language models with tools by searching on graphs.
\newblock In \emph{Computer Vision – ECCV 2024: 18th European Conference, Milan, Italy, September 29–October 4, 2024, Proceedings, Part XII}, page 89–105, Berlin, Heidelberg, 2024{\natexlab{c}}. Springer-Verlag.

\bibitem[Liu et~al.(2025)Liu, Hoang, Zhang, Zhu, Lan, Kokane, Tan, Yao, Liu, Feng, Murthy, Yang, Savarese, Niebles, Wang, Heinecke, and Xiong]{10.5555/3737916.3739641}
Zuxin Liu, Thai Hoang, Jianguo Zhang, Ming Zhu, Tian Lan, Shirley Kokane, Juntao Tan, Weiran Yao, Zhiwei Liu, Yihao Feng, Rithesh Murthy, Liangwei Yang, Silvio Savarese, Juan~Carlos Niebles, Huan Wang, Shelby Heinecke, and Caiming Xiong.
\newblock Apigen: automated pipeline for generating verifiable and diverse function-calling datasets.
\newblock In \emph{Proceedings of the 38th International Conference on Neural Information Processing Systems}, Red Hook, NY, USA, 2025. Curran Associates Inc.

\bibitem[Lu et~al.(2023)Lu, Peng, Cheng, Galley, Chang, Wu, Zhu, and Gao]{lu2023chameleon}
Pan Lu, Baolin Peng, Hao Cheng, Michel Galley, Kai-Wei Chang, Ying~Nian Wu, Song-Chun Zhu, and Jianfeng Gao.
\newblock Chameleon: Plug-and-play compositional reasoning with large language models.
\newblock In \emph{The 37th Conference on Neural Information Processing Systems (NeurIPS)}, 2023.

\bibitem[Lugmayr et~al.(2022)Lugmayr, Danelljan, Romero, Yu, Timofte, and Van~Gool]{lugmayr2022repaint}
Andreas Lugmayr, Martin Danelljan, Andres Romero, Fisher Yu, Radu Timofte, and Luc Van~Gool.
\newblock Repaint: Inpainting using denoising diffusion probabilistic models.
\newblock In \emph{Proceedings of the IEEE/CVF conference on computer vision and pattern recognition}, pages 11461--11471, 2022.

\bibitem[Mekala et~al.(2024)Mekala, Weston, Lanchantin, Raileanu, Lomeli, Shang, and Dwivedi-Yu]{mekala-etal-2024-toolverifier}
Dheeraj Mekala, Jason~E Weston, Jack Lanchantin, Roberta Raileanu, Maria Lomeli, Jingbo Shang, and Jane Dwivedi-Yu.
\newblock {TOOLVERIFIER}: Generalization to new tools via self-verification.
\newblock In \emph{Findings of the Association for Computational Linguistics: EMNLP 2024}, pages 5026--5041, Miami, Florida, USA, 2024. Association for Computational Linguistics.

\bibitem[Meng et~al.(2021)Meng, He, Song, Song, Wu, Zhu, and Ermon]{meng2021sdedit}
Chenlin Meng, Yutong He, Yang Song, Jiaming Song, Jiajun Wu, Jun-Yan Zhu, and Stefano Ermon.
\newblock Sdedit: Guided image synthesis and editing with stochastic differential equations.
\newblock \emph{arXiv preprint arXiv:2108.01073}, 2021.

\bibitem[Mokady et~al.(2023)Mokady, Hertz, Aberman, Pritch, and Cohen-Or]{mokady2023null}
Ron Mokady, Amir Hertz, Kfir Aberman, Yael Pritch, and Daniel Cohen-Or.
\newblock Null-text inversion for editing real images using guided diffusion models.
\newblock In \emph{Proceedings of the IEEE/CVF conference on computer vision and pattern recognition}, pages 6038--6047, 2023.

\bibitem[Nakano et~al.(2022)Nakano, Hilton, Balaji, Wu, Ouyang, Kim, Hesse, Jain, Kosaraju, Saunders, Jiang, Cobbe, Eloundou, Krueger, Button, Knight, Chess, and Schulman]{nakano2022webgptbrowserassistedquestionansweringhuman}
Reiichiro Nakano, Jacob Hilton, Suchir Balaji, Jeff Wu, Long Ouyang, Christina Kim, Christopher Hesse, Shantanu Jain, Vineet Kosaraju, William Saunders, Xu Jiang, Karl Cobbe, Tyna Eloundou, Gretchen Krueger, Kevin Button, Matthew Knight, Benjamin Chess, and John Schulman.
\newblock Webgpt: Browser-assisted question-answering with human feedback, 2022.

\bibitem[OpenAI(2025{\natexlab{a}})]{gpt-image}
OpenAI.
\newblock Introducing our latest image generation model in the api.
\newblock \url{https://openai.com/index/image-generation-api/}, 2025{\natexlab{a}}.

\bibitem[OpenAI(2025{\natexlab{b}})]{gpt4.1}
OpenAI.
\newblock Introducing gpt-4.1 in the api.
\newblock \url{https://openai.com/index/gpt-4-1/}, 2025{\natexlab{b}}.

\bibitem[Ouyang et~al.(2022)Ouyang, Wu, Jiang, Almeida, Wainwright, Mishkin, Zhang, Agarwal, Slama, Ray, Schulman, Hilton, Kelton, Miller, Simens, Askell, Welinder, Christiano, Leike, and Lowe]{10.5555/3600270.3602281}
Long Ouyang, Jeff Wu, Xu Jiang, Diogo Almeida, Carroll~L. Wainwright, Pamela Mishkin, Chong Zhang, Sandhini Agarwal, Katarina Slama, Alex Ray, John Schulman, Jacob Hilton, Fraser Kelton, Luke Miller, Maddie Simens, Amanda Askell, Peter Welinder, Paul Christiano, Jan Leike, and Ryan Lowe.
\newblock Training language models to follow instructions with human feedback.
\newblock In \emph{Proceedings of the 36th International Conference on Neural Information Processing Systems}, Red Hook, NY, USA, 2022. Curran Associates Inc.

\bibitem[Park et~al.(2018)Park, Lee, Yoo, and Kweon]{park2018distort}
Jongchan Park, Joon-Young Lee, Donggeun Yoo, and In~So Kweon.
\newblock Distort-and-recover: Color enhancement using deep reinforcement learning.
\newblock In \emph{Proceedings of the IEEE conference on computer vision and pattern recognition}, pages 5928--5936, 2018.

\bibitem[Patil et~al.(2024)Patil, Zhang, Wang, and Gonzalez]{patil2024gorilla}
Shishir~G Patil, Tianjun Zhang, Xin Wang, and Joseph~E. Gonzalez.
\newblock Gorilla: Large language model connected with massive {API}s.
\newblock In \emph{The Thirty-eighth Annual Conference on Neural Information Processing Systems}, 2024.

\bibitem[Qian et~al.(2024)Qian, Xiong, Liu, and Liu]{qian-etal-2024-toolink}
Cheng Qian, Chenyan Xiong, Zhenghao Liu, and Zhiyuan Liu.
\newblock Toolink: Linking toolkit creation and using through chain-of-solving on open-source model.
\newblock In \emph{Proceedings of the 2024 Conference of the North American Chapter of the Association for Computational Linguistics: Human Language Technologies (Volume 1: Long Papers)}, pages 831--854, Mexico City, Mexico, 2024. Association for Computational Linguistics.

\bibitem[Qin et~al.(2024)Qin, Liang, Ye, Zhu, Yan, Lu, Lin, Cong, Tang, Qian, Zhao, Hong, Tian, Xie, Zhou, Gerstein, dahai li, Liu, and Sun]{qin2024toolllm}
Yujia Qin, Shihao Liang, Yining Ye, Kunlun Zhu, Lan Yan, Yaxi Lu, Yankai Lin, Xin Cong, Xiangru Tang, Bill Qian, Sihan Zhao, Lauren Hong, Runchu Tian, Ruobing Xie, Jie Zhou, Mark Gerstein, dahai li, Zhiyuan Liu, and Maosong Sun.
\newblock Tool{LLM}: Facilitating large language models to master 16000+ real-world {API}s.
\newblock In \emph{The Twelfth International Conference on Learning Representations}, 2024.

\bibitem[Qu et~al.(2024)Qu, Dai, Wei, Cai, Wang, Yin, Xu, and Wen]{10.1145/3627673.3679847}
Changle Qu, Sunhao Dai, Xiaochi Wei, Hengyi Cai, Shuaiqiang Wang, Dawei Yin, Jun Xu, and Ji-Rong Wen.
\newblock Towards completeness-oriented tool retrieval for large language models.
\newblock In \emph{Proceedings of the 33rd ACM International Conference on Information and Knowledge Management}, page 1930–1940, New York, NY, USA, 2024. Association for Computing Machinery.

\bibitem[Qu et~al.(2025)Qu, Dai, Wei, Cai, Wang, Yin, Xu, and Wen]{qu2025from}
Changle Qu, Sunhao Dai, Xiaochi Wei, Hengyi Cai, Shuaiqiang Wang, Dawei Yin, Jun Xu, and Ji-Rong Wen.
\newblock From exploration to mastery: Enabling {LLM}s to master tools via self-driven interactions.
\newblock In \emph{The Thirteenth International Conference on Learning Representations}, 2025.

\bibitem[Radford et~al.(2021)Radford, Kim, Hallacy, Ramesh, Goh, Agarwal, Sastry, Askell, Mishkin, Clark, et~al.]{radford2021learning}
Alec Radford, Jong~Wook Kim, Chris Hallacy, Aditya Ramesh, Gabriel Goh, Sandhini Agarwal, Girish Sastry, Amanda Askell, Pamela Mishkin, Jack Clark, et~al.
\newblock Learning transferable visual models from natural language supervision.
\newblock In \emph{International conference on machine learning}, pages 8748--8763. PmLR, 2021.

\bibitem[Ramesh et~al.(2022)Ramesh, Dhariwal, Nichol, Chu, and Chen]{ramesh2022hierarchical}
Aditya Ramesh, Prafulla Dhariwal, Alex Nichol, Casey Chu, and Mark Chen.
\newblock Hierarchical text-conditional image generation with clip latents.
\newblock \emph{arXiv preprint arXiv:2204.06125}, 1\penalty0 (2):\penalty0 3, 2022.

\bibitem[Reuss et~al.(2024)Reuss, Ya{\u{g}}murlu, Wenzel, and Lioutikov]{reuss2024multimodal}
Moritz Reuss, {\"O}mer~Erdin{\c{c}} Ya{\u{g}}murlu, Fabian Wenzel, and Rudolf Lioutikov.
\newblock Multimodal diffusion transformer: Learning versatile behavior from multimodal goals.
\newblock In \emph{Robotics: Science and Systems}, 2024.

\bibitem[Rombach et~al.(2021)Rombach, Blattmann, Lorenz, Esser, and Ommer]{rombach2021highresolution}
Robin Rombach, Andreas Blattmann, Dominik Lorenz, Patrick Esser, and Björn Ommer.
\newblock High-resolution image synthesis with latent diffusion models, 2021.

\bibitem[Schick et~al.(2023)Schick, Dwivedi-Yu, Dessi, Raileanu, Lomeli, Hambro, Zettlemoyer, Cancedda, and Scialom]{schick2023toolformer}
Timo Schick, Jane Dwivedi-Yu, Roberto Dessi, Roberta Raileanu, Maria Lomeli, Eric Hambro, Luke Zettlemoyer, Nicola Cancedda, and Thomas Scialom.
\newblock Toolformer: Language models can teach themselves to use tools.
\newblock In \emph{Thirty-seventh Conference on Neural Information Processing Systems}, 2023.

\bibitem[Shao et~al.(2024)Shao, Wang, Zhu, Xu, Song, Bi, Zhang, Zhang, Li, Wu, et~al.]{shao2024deepseekmath}
Zhihong Shao, Peiyi Wang, Qihao Zhu, Runxin Xu, Junxiao Song, Xiao Bi, Haowei Zhang, Mingchuan Zhang, YK Li, Yang Wu, et~al.
\newblock Deepseekmath: Pushing the limits of mathematical reasoning in open language models.
\newblock \emph{arXiv preprint arXiv:2402.03300}, 2024.

\bibitem[Shen et~al.(2023)Shen, Song, Tan, Li, Lu, and Zhuang]{10.5555/3666122.3667779}
Yongliang Shen, Kaitao Song, Xu Tan, Dongsheng Li, Weiming Lu, and Yueting Zhuang.
\newblock Hugginggpt: solving ai tasks with chatgpt and its friends in hugging face.
\newblock In \emph{Proceedings of the 37th International Conference on Neural Information Processing Systems}, Red Hook, NY, USA, 2023. Curran Associates Inc.

\bibitem[Sheng et~al.(2024)Sheng, Zhang, Ye, Wu, Zhang, Zhang, Peng, Lin, and Wu]{sheng2024hybridflow}
Guangming Sheng, Chi Zhang, Zilingfeng Ye, Xibin Wu, Wang Zhang, Ru Zhang, Yanghua Peng, Haibin Lin, and Chuan Wu.
\newblock Hybridflow: A flexible and efficient rlhf framework.
\newblock \emph{arXiv preprint arXiv: 2409.19256}, 2024.

\bibitem[Shi et~al.(2020)Shi, Xu, Bui, Dernoncourt, Wen, and Xu]{shi2020benchmark}
Jing Shi, Ning Xu, Trung Bui, Franck Dernoncourt, Zheng Wen, and Chenliang Xu.
\newblock A benchmark and baseline for language-driven image editing.
\newblock In \emph{Proceedings of the Asian Conference on Computer Vision}, 2020.

\bibitem[Shi et~al.(2021)Shi, Xu, Xu, Bui, Dernoncourt, and Xu]{gier}
Jing Shi, Ning Xu, Yihang Xu, Trung Bui, Franck Dernoncourt, and Chenliang Xu.
\newblock Learning by planning: Language-guided global image editing.
\newblock In \emph{Proceedings of the IEEE/CVF Conference on Computer Vision and Pattern Recognition}, pages 13590--13599, 2021.

\bibitem[Shi et~al.(2022)Shi, Xu, Zheng, Smith, Luo, and Xu]{Shi_2022_CVPR}
Jing Shi, Ning Xu, Haitian Zheng, Alex Smith, Jiebo Luo, and Chenliang Xu.
\newblock Spaceedit: Learning a unified editing space for open-domain image color editing.
\newblock In \emph{Proceedings of the IEEE/CVF Conference on Computer Vision and Pattern Recognition (CVPR)}, pages 19730--19739, 2022.

\bibitem[Shi et~al.(2024)Shi, Gao, Chen, Feng, Yan, Shi, Yin, Ren, Verberne, and Ren]{shi-etal-2024-learning}
Zhengliang Shi, Shen Gao, Xiuyi Chen, Yue Feng, Lingyong Yan, Haibo Shi, Dawei Yin, Pengjie Ren, Suzan Verberne, and Zhaochun Ren.
\newblock Learning to use tools via cooperative and interactive agents.
\newblock In \emph{Findings of the Association for Computational Linguistics: EMNLP 2024}, pages 10642--10657, Miami, Florida, USA, 2024. Association for Computational Linguistics.

\bibitem[Sun et~al.(2022)Sun, Chen, Chen, Dai, Zhu, and Yu]{sun-etal-2022-meta}
Liangtai Sun, Xingyu Chen, Lu Chen, Tianle Dai, Zichen Zhu, and Kai Yu.
\newblock {META}-{GUI}: Towards multi-modal conversational agents on mobile {GUI}.
\newblock In \emph{Proceedings of the 2022 Conference on Empirical Methods in Natural Language Processing}, pages 6699--6712, Abu Dhabi, United Arab Emirates, 2022. Association for Computational Linguistics.

\bibitem[Sur{\'\i}s et~al.(2023)Sur{\'\i}s, Menon, and Vondrick]{Suris_2023_ICCV}
D{\'\i}dac Sur{\'\i}s, Sachit Menon, and Carl Vondrick.
\newblock Vipergpt: Visual inference via python execution for reasoning.
\newblock In \emph{Proceedings of the IEEE/CVF International Conference on Computer Vision (ICCV)}, pages 11888--11898, 2023.

\bibitem[Team(2025)]{qwen3technicalreport}
Qwen Team.
\newblock Qwen3 technical report, 2025.

\bibitem[Wang et~al.(2023)Wang, Xie, Jiang, Mandlekar, Xiao, Zhu, Fan, and Anandkumar]{wang2023voyager}
Guanzhi Wang, Yuqi Xie, Yunfan Jiang, Ajay Mandlekar, Chaowei Xiao, Yuke Zhu, Linxi Fan, and Anima Anandkumar.
\newblock Voyager: An open-ended embodied agent with large language models.
\newblock \emph{arXiv preprint arXiv: Arxiv-2305.16291}, 2023.

\bibitem[Wang et~al.(2024)Wang, Li, Li, and Liu]{wang2024genartist}
Zhenyu Wang, Aoxue Li, Zhenguo Li, and Xihui Liu.
\newblock Gen{A}rtist: Multimodal llm as an agent for unified image generation and editing.
\newblock \emph{Advances in Neural Information Processing Systems}, 37:\penalty0 128374--128395, 2024.

\bibitem[Wu et~al.(2023)Wu, Yin, Qi, Wang, Tang, and Duan]{wu2023visualchatgpttalkingdrawing}
Chenfei Wu, Shengming Yin, Weizhen Qi, Xiaodong Wang, Zecheng Tang, and Nan Duan.
\newblock Visual chatgpt: Talking, drawing and editing with visual foundation models, 2023.

\bibitem[Wu et~al.(2025)Wu, Li, Zhou, Lin, Gao, Yan, Yin, Bai, Xu, Chen, et~al.]{wu2025qwen}
Chenfei Wu, Jiahao Li, Jingren Zhou, Junyang Lin, Kaiyuan Gao, Kun Yan, Sheng-ming Yin, Shuai Bai, Xiao Xu, Yilei Chen, et~al.
\newblock Qwen-image technical report.
\newblock \emph{arXiv preprint arXiv:2508.02324}, 2025.

\bibitem[Xu et~al.(2025{\natexlab{a}})Xu, Wang, Zhu, Pan, Chen, Chen, and Yu]{xu2025alignmentefficienttoolcalling}
Hongshen Xu, Zihan Wang, Zichen Zhu, Lei Pan, Xingyu Chen, Lu Chen, and Kai Yu.
\newblock Alignment for efficient tool calling of large language models, 2025{\natexlab{a}}.

\bibitem[Xu et~al.(2025{\natexlab{b}})Xu, yang, Zhu, Lan, Wang, Wu, Ji, Chen, Fung, and Yu]{xu2025delusionslargelanguagemodels}
Hongshen Xu, Zixv yang, Zichen Zhu, Kunyao Lan, Zihan Wang, Mengyue Wu, Ziwei Ji, Lu Chen, Pascale Fung, and Kai Yu.
\newblock Delusions of large language models, 2025{\natexlab{b}}.

\bibitem[Xu et~al.(2025{\natexlab{c}})Xu, Zhu, Pan, Wang, Zhu, Ma, Cao, Chen, and Yu]{xu2025reducing}
Hongshen Xu, Zichen Zhu, Lei Pan, Zihan Wang, Su Zhu, Da Ma, Ruisheng Cao, Lu Chen, and Kai Yu.
\newblock Reducing tool hallucination via reliability alignment.
\newblock In \emph{Forty-second International Conference on Machine Learning}, 2025{\natexlab{c}}.

\bibitem[Xu et~al.(2024)Xu, Huang, Pan, Ma, and Chai]{Xu_2024_CVPR}
Sihan Xu, Yidong Huang, Jiayi Pan, Ziqiao Ma, and Joyce Chai.
\newblock Inversion-free image editing with language-guided diffusion models.
\newblock In \emph{Proceedings of the IEEE/CVF Conference on Computer Vision and Pattern Recognition (CVPR)}, pages 9452--9461, 2024.

\bibitem[Yang et~al.(2023)Yang, Gu, Zhang, Zhang, Chen, Sun, Chen, and Wen]{Yang_2023_CVPR}
Binxin Yang, Shuyang Gu, Bo Zhang, Ting Zhang, Xuejin Chen, Xiaoyan Sun, Dong Chen, and Fang Wen.
\newblock Paint by example: Exemplar-based image editing with diffusion models.
\newblock In \emph{Proceedings of the IEEE/CVF Conference on Computer Vision and Pattern Recognition (CVPR)}, pages 18381--18391, 2023.

\bibitem[Yao et~al.(2023)Yao, Zhao, Yu, Du, Shafran, Narasimhan, and Cao]{yao2023react}
Shunyu Yao, Jeffrey Zhao, Dian Yu, Nan Du, Izhak Shafran, Karthik~R Narasimhan, and Yuan Cao.
\newblock React: Synergizing reasoning and acting in language models.
\newblock In \emph{The Eleventh International Conference on Learning Representations}, 2023.

\bibitem[Yuan et~al.(2024)Yuan, Chen, Wang, Fung, Peng, and Ji]{DBLP:conf/iclr/YuanC000J24}
Lifan Yuan, Yangyi Chen, Xingyao Wang, Yi Fung, Hao Peng, and Heng Ji.
\newblock Craft: Customizing llms by creating and retrieving from specialized toolsets.
\newblock In \emph{ICLR}, 2024.

\bibitem[Zeng et~al.(2023)Zeng, Attarian, brian ichter, Choromanski, Wong, Welker, Tombari, Purohit, Ryoo, Sindhwani, Lee, Vanhoucke, and Florence]{zeng2023socratic}
Andy Zeng, Maria Attarian, brian ichter, Krzysztof~Marcin Choromanski, Adrian Wong, Stefan Welker, Federico Tombari, Aveek Purohit, Michael~S Ryoo, Vikas Sindhwani, Johnny Lee, Vincent Vanhoucke, and Pete Florence.
\newblock Socratic models: Composing zero-shot multimodal reasoning with language.
\newblock In \emph{The Eleventh International Conference on Learning Representations}, 2023.

\bibitem[Zhang et~al.(2024)Zhang, Li, He, Zhang, Qiao, Qin, Ma, Kang, Lin, Rajmohan, Zhang, and Zhang]{zhang2024ufo}
Chaoyun Zhang, Liqun Li, Shilin He, Xu Zhang, Bo Qiao, Si Qin, Minghua Ma, Yu Kang, Qingwei Lin, Saravan Rajmohan, Dongmei Zhang, and Qi Zhang.
\newblock {UFO: A UI-Focused Agent for Windows OS Interaction}.
\newblock \emph{arXiv preprint arXiv:2402.07939}, 2024.

\bibitem[Zhang et~al.(2023)Zhang, Mo, Chen, Sun, and Su]{zhang2023magicbrush}
Kai Zhang, Lingbo Mo, Wenhu Chen, Huan Sun, and Yu Su.
\newblock Magicbrush: A manually annotated dataset for instruction-guided image editing.
\newblock \emph{Advances in Neural Information Processing Systems}, 36:\penalty0 31428--31449, 2023.

\bibitem[Zhang et~al.(2025)Zhang, Rossi, Kveton, Shao, Yang, Zamani, Dernoncourt, Barrow, Yu, Kim, Zhang, Gu, Derr, Chen, Wu, Chen, Wang, Mitra, Lipka, Ahmed, and Wang]{zhang2025personalizationlargelanguagemodels}
Zhehao Zhang, Ryan~A. Rossi, Branislav Kveton, Yijia Shao, Diyi Yang, Hamed Zamani, Franck Dernoncourt, Joe Barrow, Tong Yu, Sungchul Kim, Ruiyi Zhang, Jiuxiang Gu, Tyler Derr, Hongjie Chen, Junda Wu, Xiang Chen, Zichao Wang, Subrata Mitra, Nedim Lipka, Nesreen Ahmed, and Yu Wang.
\newblock Personalization of large language models: A survey, 2025.

\bibitem[Zhao et~al.(2024)Zhao, Ma, Chen, Si, Wu, An, Yu, Zhang, Li, and Chang]{NEURIPS2024_05a30a0f}
Haozhe Zhao, Xiaojian Ma, Liang Chen, Shuzheng Si, Rujie Wu, Kaikai An, Peiyu Yu, Minjia Zhang, Qing Li, and Baobao Chang.
\newblock Ultraedit: Instruction-based fine-grained image editing at scale.
\newblock In \emph{Advances in Neural Information Processing Systems}, pages 3058--3093. Curran Associates, Inc., 2024.

\bibitem[Zheng et~al.(2024)Zheng, Li, Liu, Liu, Luan, and Wang]{zheng-etal-2024-toolrerank}
Yuanhang Zheng, Peng Li, Wei Liu, Yang Liu, Jian Luan, and Bin Wang.
\newblock {T}ool{R}erank: Adaptive and hierarchy-aware reranking for tool retrieval.
\newblock In \emph{Proceedings of the 2024 Joint International Conference on Computational Linguistics, Language Resources and Evaluation (LREC-COLING 2024)}, pages 16263--16273, Torino, Italia, 2024. ELRA and ICCL.

\bibitem[Zhu et~al.(2025)Zhu, Tang, Li, Liu, Xu, Lan, Zhang, Jiang, Zhou, Wang, Zhang, Sun, Wang, Sun, Chen, and Yu]{zhu-etal-2025-moba}
Zichen Zhu, Hao Tang, Yansi Li, Dingye Liu, Hongshen Xu, Kunyao Lan, Danyang Zhang, Yixuan Jiang, Hao Zhou, Chenrun Wang, Situo Zhang, Liangtai Sun, Yixiao Wang, Yuheng Sun, Lu Chen, and Kai Yu.
\newblock {M}ob{A}: Multifaceted memory-enhanced adaptive planning for efficient mobile task automation.
\newblock In \emph{Proceedings of the 2025 Conference of the Nations of the Americas Chapter of the Association for Computational Linguistics: Human Language Technologies (System Demonstrations)}, pages 535--549, Albuquerque, New Mexico, 2025. Association for Computational Linguistics.

\bibitem[Zhuang et~al.(2024)Zhuang, Chen, Yu, Mitra, Bursztyn, Rossi, Sarkhel, and Zhang]{DBLP:conf/iclr/ZhuangC0MBRS024}
Yuchen Zhuang, Xiang Chen, Tong Yu, Saayan Mitra, Victor Bursztyn, Ryan~A. Rossi, Somdeb Sarkhel, and Chao Zhang.
\newblock Toolchain*: Efficient action space navigation in large language models with a* search.
\newblock In \emph{ICLR}, 2024.

\end{thebibliography}
}

\clearpage
\appendix

\section{Future Work and Limitations}
Our study is limited by (i) a restricted editing space with only 16 global operations, lacking local/semantic controls; (ii) a simplified \texttt{matplotlib}-based engine that cannot fully reproduce professional rendering pipelines, creating a sim-to-real gap; (iii) modest data scale and partial reliance on synthetic/pseudo-labeled supervision; (iv) template-driven construction of Image-Refine that may not capture authentic, nuanced user corrections; and (v) evaluation that leans on pixel distances, an internal reward model, and a small user study, which together under-represent subjective aesthetics and long-term satisfaction.

Future work will expand the action space to local and semantic-aware tools (e.g., segmentation-guided or generative-assisted operations) while preserving realism; build a higher-fidelity, Linux-compatible editing backend to reduce latency and deployment mismatch; collect larger and more diverse, human-in-the-loop preference/refinement data; and adopt stronger evaluations via broader, longitudinal user studies and perceptual-quality metrics.

\newpage
\section{List of Notations}

We list the notations of symbols used in this paper.

\begin{table}[h]
\centering
\caption{Key symbols and their meanings used in the main paper.}
\label{tab:notation}
\setlength{\tabcolsep}{5pt}
\small
\begin{tabular}{c p{6.5cm}}
\toprule
\textbf{Symbol} & \textbf{Meaning} \\
\midrule
$\mathcal{E}$ & Simulation image editor. \\
$Q$ & User instruction/query text. \\
$t$ & A single tool call with corresponding parameter. \\
$T$ & A set of tool parameters (one parameter per tool). $T=\{t_1,\dots,t_m\}$  \\
$T\setminus\{t\}$ & Tool set with tool $t$ removed. \\
$\mathcal{M}$ & \agent{} which maps multimodal inputs to $T$ (edit/refine) or $Q$ (summary). \\
$I_{\text{ori}}$ & Original unedited image. \\
$I_{\text{ref}}$ & Reference image retouched by human expert. \\
$I_{\text{edit}}$ & Edited image after applying $T$ via the editor $\mathcal{E}$. \\
$I_{\text{his}}$ & User-edited historical image. \\
$\mathcal{L}(I_a,I_b)$ & Pixel-level distance between images (mean of L1 and L2). \\
$L$ & Shorthand for $\mathcal{L}(I_{\text{edit}},\,I_{\text{ref}})$. \\
$R_L$ & Likeness Improvement Reward. \\
$R_U$ & Tool Usefulness Reward. \\
$R_A$ & Summary Alignment Reward. \\
$\mathds{1}(\cdot)$ & Indicator function (1 if condition holds, else 0). \\
\bottomrule
\end{tabular}
\end{table}

\newpage
\section{List of Available Tools}
We list all 16 image editing tools used in this paper.
\begin{table*}[t]
\centering
\caption{Available image editing tools in our simulation image editor.}
\label{tab:editing_tools}
\small
\begin{tabularx}{\textwidth}{@{}>{\hsize=0.7\hsize}X>{\hsize=1.5\hsize}X>{\hsize=0.8\hsize}X@{}}
\toprule
\textbf{Function} & \textbf{Description} & \textbf{Example of Value} \\ \midrule
exposure & Adjusts the overall image exposure & 30 (brighter) \\
brightness & Adjusts overall image brightness & 30 (brighter) \\
contrast & Adjusts the difference between light and dark areas & 40 (higher contrast) \\
natural\_contrast & Adjusts natural contrast & 40 (higher contrast) \\
highlights & Adjusts the brightest areas of the image & -50 (darker highlights) \\
shadows & Adjusts the darkest areas of the image & 50 (lighter shadows) \\
whites & Adjusts the white point of the image & -20 (duller whites) \\
blacks & Adjusts the black point of the image & 20 (lighter blacks) \\
saturation & Adjusts the color intensity & -100 (black \& white) \\
vibrance & Boosts muted colors more than saturated colors & 50 (more vibrant) \\
temperature & Adjusts the color temperature (warm/cool) & -20 (cooler) \\
tint & Adjusts the color tint (green/magenta shift) & 50 (more green) \\
sharpness & Adjusts the clarity of edges & 80 (sharper) \\
vignette & Adds a dark or bright effect to the corners & -30 (darker corners) \\
fade & Applies a washed-out look to the image & 60 (more faded) \\
grain & Adds film grain or noise to the image & 20 (more grain) \\ 
\bottomrule
\end{tabularx}%
\end{table*}

\clearpage
\section{Greedy Tool-wise Search Algorithm}
\label{app:greedy-search}

We provide a detailed pseudo-code for the algorithm we used during optimal tool search.
\begin{algorithm}[h]
\caption{Greedy Tool-wise Parameter Search.} 
\DontPrintSemicolon
\SetAlgoLined
\SetKwInOut{Input}{Input}
\SetKwInOut{Output}{Output}
\SetKwFunction{Loss}{Loss}

\vspace{0.2cm}
\Input{$I_{\mathrm{ori}},\, I_{\mathrm{ref}},\, E$ (editor); initial $T$ (dict; tools $\mapsto$ values in $[-100,100]$)\\
\hspace{1.85em}Offset set $\Delta=\{\pm 50,\pm 25,\pm 10,\pm 5\}$, threshold $\tau>0$}
\Output{Refined tool-call $T'$}

\BlankLine
\textbf{Editing call:} $I_{\mathrm{edit}} \gets E(I_{\mathrm{ori}}, T_{\mathrm{edit}})$\;
\textbf{Distance:} $\mathcal{L}(I_{\mathrm{edit}}, I_{\mathrm{ref}})$\;

\BlankLine
\SetKwProg{Fn}{Function}{:}{end}
\Fn{\Loss{$T$}}{
  $I_{\mathrm{edit}} \gets E(I_{\mathrm{ori}}, T)$\;
  \Return{$\mathcal{L}(I_{\mathrm{edit}}, I_{\mathrm{ref}})$}\;
}

$L^\star \gets \Loss{T}$;\quad $S \gets \mathrm{keys}(T)$\;
\While{$S \neq \emptyset$}{
  $(t^\star,\delta^\star,\mathrm{gain}^\star) \gets (\varnothing, 0, 0)$\;
  \ForEach{$t \in S$}{
    \ForEach{$\delta \in \Delta$}{
      $T' \gets T$;\quad $T'[t] \gets \mathrm{clip}\big(T[t]+\delta,-100,100\big)$\;
      $L' \gets \Loss{T'}$;\quad $\mathrm{gain} \gets L^\star - L'$\;
      \If{$\mathrm{gain} > \mathrm{gain}^\star$}{
        $(t^\star,\delta^\star,\mathrm{gain}^\star) \gets (t,\delta,\mathrm{gain})$\;
      }
    }
  }
  \If{$\mathrm{gain}^\star \le \tau$}{
    \textbf{break} \tcp*{no sufficient improvement}
  }
  $T[t^\star] \gets \mathrm{clip}\big(T[t^\star]+\delta^\star,-100,100\big)$\;
  $L^\star \gets L^\star - \mathrm{gain}^\star$;\quad $S \gets S \setminus \{t^\star\}$\;
}
\Return{$T$} \tcp*{$T$ is the refined $T'$}
\end{algorithm}

\newpage
\section{Ablations on Parameter Search.}
\begin{figure}[h]
\centering
\includegraphics[width=1\linewidth,trim=0 0 0 0,clip]{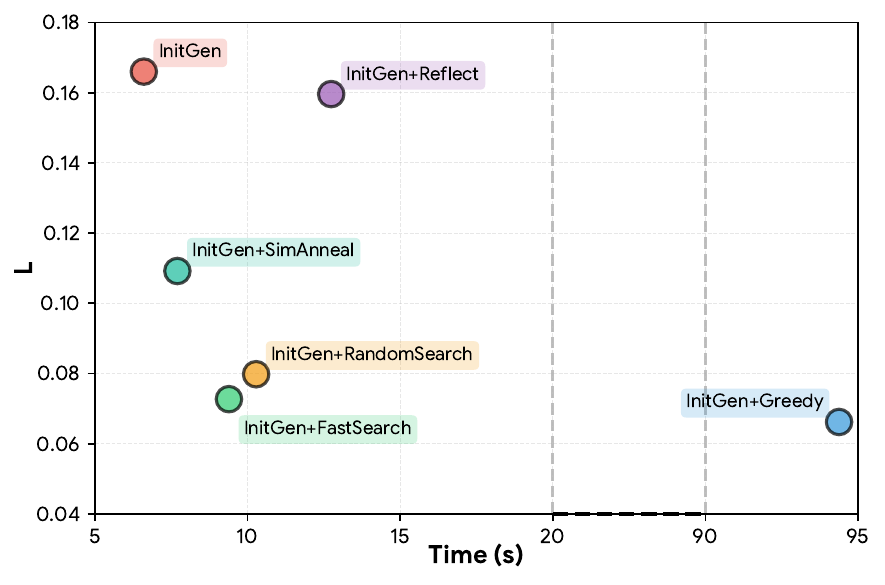} 
\caption{Comparison of parameter search algorithms. Closer to left-bottom is better.}
  \label{fig:search-comparison}
\end{figure}

We compare six search methods during parameter search process: (1) \textit{InitGen}, the initial attempt from \texttt{GPT-4.1}; (2) \textit{+Reflect}, a second search round with reflection; (3) \textit{+SimAnneal}, simulated annealing with 1000 iterations; (4) \textit{+RandomSearch}, random perturbation with $\Delta\in[-100,100]$ for 8 attempts; (5) \textit{+FastSearch}, our proposed algorithm; and (6) \textit{+Greedy}, exhaustive testing of each parameter from $-100$ to $+100$. As shown in \Cref{fig:search-comparison}, our search strategy provides a favorable balance between search cost and final quality.

\clearpage
\section{Data Detail}
We provide the detailed training data composition of all three stages.

\begin{table*}[t]
\centering
\caption{Composition of training data across stages. Stage columns are the sampling ratio used.}
\label{tab:data_mix_by_stage}
\small
\begin{tabularx}{\textwidth}{@{} >{\hsize=1.2\hsize}X >{\hsize=1.4\hsize}X r*{3}{>{\centering\arraybackslash\hsize=0.8\hsize}X} @{}}
\toprule
\multirow{2.5}{*}{\textbf{Image Source}} & \multirow{2.5}{*}{\textbf{Task}} & \multirow{2}{*}{\textbf{\# Items}} &
\multicolumn{3}{c}{\textbf{Sampling Ratio per Stage}} \\
\cmidrule(lr){4-6}
 &  &  & \textbf{Stage 1} & \textbf{Stage 2} & \textbf{Stage 3} \\
\midrule
\multirow{5}{*}{GIER~\cite{gier}}
& Image-Edit                 & 25{,}763  & 1 & 1 & 2 \\
& Image-Summary              & 3{,}583   & 1 & 1 & 5 \\
& Image-Edit-Synthesis       & 173{,}802 &   &   & 0.2 \\
& Image-Summary-Synthesis    & 17{,}110  &   &   & 1 \\
& Image-Refine-Synthesis     & 7{,}166   &   &   & 2 \\
\midrule
\multirow{3}{*}{MIT--Adobe FiveK~\cite{fivek}}
& Image-Edit-Synthesis       & 174{,}006 &   &   & 0.2 \\
& Image-Summary-Synthesis    & 17{,}102  &   &   & 1 \\
& Image-Refine-Synthesis     & 7{,}136   &   &   & 2 \\
\bottomrule
\end{tabularx}
\end{table*}

\clearpage
\section{Reward Model Setup}
\label{sec:rm-setup}

We fine-tune a lightweight judge, \texttt{Qwen3-0.6B}\cite{qwen3technicalreport}, to evaluate the consistency between a predicted preference $Q_{\text{pred}}$ and a ground-truth preference $Q_{\text{ref}}$. The RM outputs $R_A\in[-10,10]$ based on semantic alignment, attribute coverage, and specificity. We train with a batch size of 256, learning rate of $5{\times}10^{-5}$, for 10 epochs on $\sim$140k SFT items (mixture of synthetic and real), totaling $\sim$5.5k steps. On a 3k-sample test split, we report mean absolute error (MAE) and accuracy as metrics.

\begin{table}[h]
\centering
\caption{Results of reward model. MAE $\downarrow$ is absolute error on $[-10,10]$; Acc $\uparrow$ is the fraction of accurate judgments.}
\label{tab:rm-results}
\setlength{\tabcolsep}{8pt}
\small
\begin{tabular}{@{}lcc@{}}
\toprule
\textbf{Model} & \textbf{MAE} (↓) & \textbf{Acc} (↑) \\
\midrule
GPT 4.1 mini\cite{gpt4.1} & 5.0013 & 0.1420 \\
Gemini 2.5 Flash-Lite\cite{gemini2.5} & 5.1923 & 0.0983 \\
Qwen3-0.6B & 7.2412 & 0.0742 \\
\midrule
IEA-Summary-RM & 1.1381 & 0.6654 \\
\bottomrule
\end{tabular}
\end{table}

\begin{figure}[h]
\centering
\includegraphics[width=1\linewidth,trim=0 0 0 0,clip]{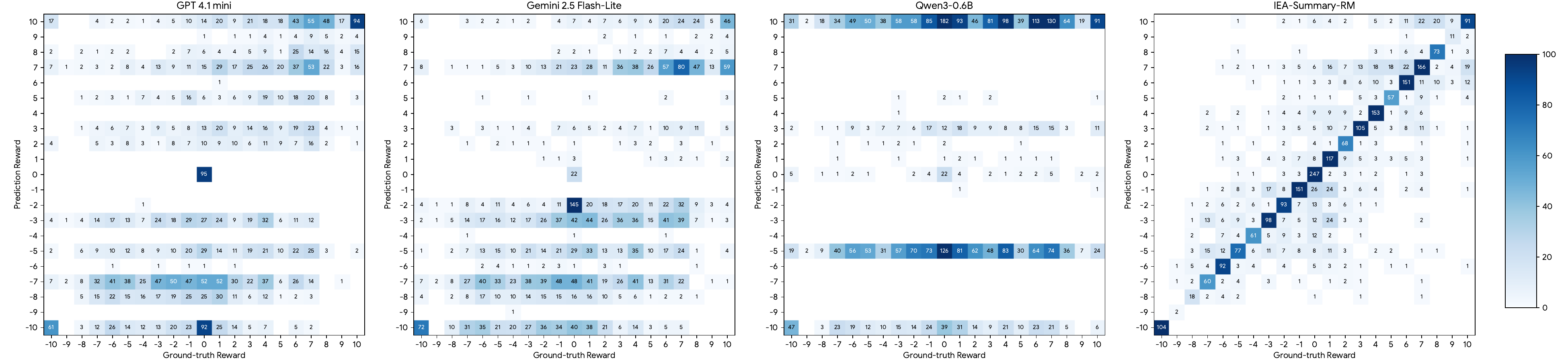} 
\caption{Distribution of predicted reward vs.\ ground-truth scores. }
\label{fig:rm_heatmap}
\end{figure}

Qualitatively, the SFT RM sharpens discrimination among near-miss summaries (e.g., “brighter but less colorful” vs.\ “brighter and more colorful”), while penalizing vacuous or off-topic outputs. We visualize predicted vs.\ ground-truth scores as a 2D histogram in \Cref{fig:rm_heatmap}. IEA-Summary-RM exhibits tight mass along the diagonal, with heavier tails primarily at extreme scores.

\section{GUI Interface}
\begin{figure}[h]
\centering
\includegraphics[width=\linewidth]{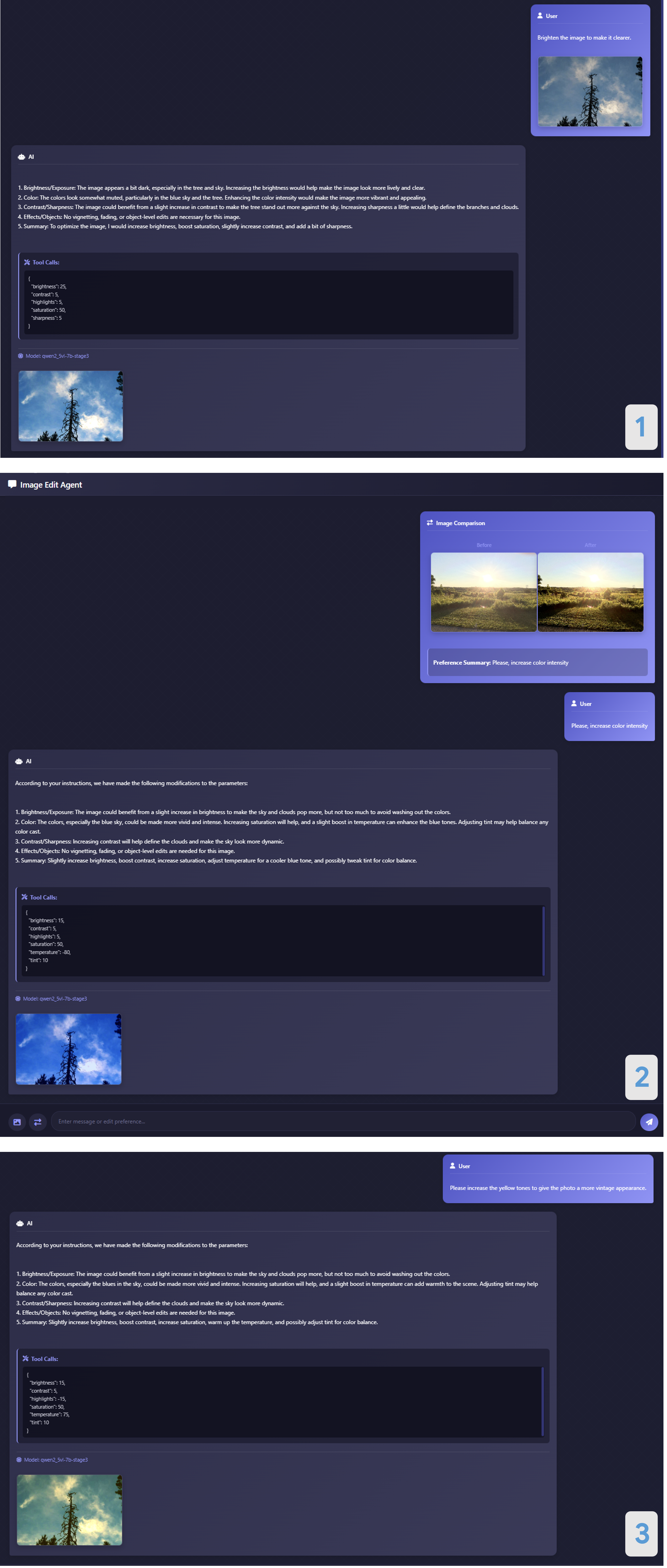}  
\caption{
        Interface of \textbf{I}mage \textbf{E}diting \textbf{A}gent.
}
\label{fig:GUI_interface}
\end{figure}
To better domonstrate the functions of \agent{}, we build a online demo that supports:
\textbf{(1)} The user inputs an original image and puts forward requirements: \textit{“Brighten the image to make it clearer.”}
\textbf{(2)} The user provides two groups of historically edited images, and \agent{} summarizes the style instructions: \textit{“Please, increase color intensity.”}
\textbf{(3)} Based on the previous context, the user puts forward new requirements: \textit{Please increase the yellow tones to give the photo a more vintage appearance.}
Users can input commands via natural language, and the \agent{} will call parameterized tools for editing and present the results.

\clearpage

\begin{table*}[t]
\centering
\caption{Expanded results on the 50 user-study samples. Rank(A/B) are average human rankings for instruction following and image quality, respectively.}
\label{tab:user-study-extended}
\small
\renewcommand{\arraystretch}{1}
\begin{tabular}{lcccccccc}
\toprule
\textbf{Model} &
\makecell{\textbf{Rank}\\\textbf{(A$\downarrow$)}} &
\makecell{\textbf{Rank}\\\textbf{(B$\downarrow$)}} &
\makecell{\textbf{L}\\\textbf{($\downarrow$)}} &
\makecell{\textbf{$R_L$}\\\textbf{($\uparrow$)}} &
\makecell{\textbf{CLIP}\\\textbf{($\uparrow$)}} &
\makecell{\textbf{LLM}\\\textbf{(Ins$\uparrow$)}} &
\makecell{\textbf{LLM}\\\textbf{(Sim$\uparrow$)}} &
\makecell{\textbf{LLM}\\\textbf{(Qua$\uparrow$)}} \\
\midrule
Reference & 2.91 & 2.87 & 0.00 & 1.00 & 1.00 & 9.38 & 9.46 & 5.72 \\
Origin & - & - & 0.17 & 0.00 & 0.94 & 8.94 & 3.22 & 4.78 \\
\midrule
GPT-Image-1 & \textcolor{DDG}{2.93} & \textcolor{MG}{4.15} & \textcolor{MR}{0.22} & \textcolor{MR}{-0.34} & \textcolor{MR}{0.91} & \textcolor{DDG}{9.60} & \textcolor{DDG}{8.78} & 5.50 \\
Qwen-Image-Edit & \textcolor{DG}{3.26} & \textcolor{DG}{3.74} & \textcolor{DG}{0.18} & \textcolor{DG}{-0.13} & \textcolor{MR}{0.91} & \textcolor{DG}{9.58} & \textcolor{DG}{7.54} & \textcolor{DDG}{5.76} \\
PatchDPO & - & - & \textcolor{DDR}{0.27} & \textcolor{DDR}{-0.60} & \textcolor{DDR}{0.85} & 8.96 & \textcolor{DDR}{4.36} & \textcolor{MR}{5.18} \\
GenArtist & - & - & \textcolor{MG}{0.19} & \textcolor{MG}{-0.16} & \textcolor{MG}{0.92} & 9.15 & \textcolor{DR}{4.54} & \textcolor{DDR}{4.96} \\
JarvisArt & \textcolor{DDR}{6.13} & \textcolor{MR}{5.36} & \textcolor{MR}{0.22} & -0.33 & \textcolor{MR}{0.91} & \textcolor{MG}{9.56} & 4.78 & \textcolor{DG}{5.70} \\
GPT-4.1 & \textcolor{DR}{5.86} & \textcolor{DDR}{5.84} & 0.21 & -0.31 & \textcolor{MR}{0.91} & \textcolor{DR}{8.24} & 4.80 & 5.22 \\
Gemini-2.5-Pro & \textcolor{MR}{5.47} & \textcolor{DR}{5.79} & \textcolor{DR}{0.23} & \textcolor{DR}{-0.42} & \textcolor{DR}{0.88} & \textcolor{DDR}{7.62} & \textcolor{MR}{4.74} & 5.32 \\
Qwen2.5-VL-7B & 4.80 & 4.55 & 0.20 & -0.22 & \textcolor{DG}{0.93} & \textcolor{MR}{8.30} & 5.62 & \textcolor{DR}{5.12} \\
\midrule
Ours & \textcolor{MG}{4.64} & \textcolor{DDG}{3.69} & \textcolor{DDG}{0.13} & \textcolor{DDG}{0.14} & \textcolor{DDG}{0.94} & 9.40 & \textcolor{MG}{6.18} & \textcolor{MG}{5.62} \\
\bottomrule
\end{tabular}
\end{table*}

\section{User Study Detail}
\begin{figure}[h]
\centering
\includegraphics[width=\linewidth,trim=0 65 0 85,clip]{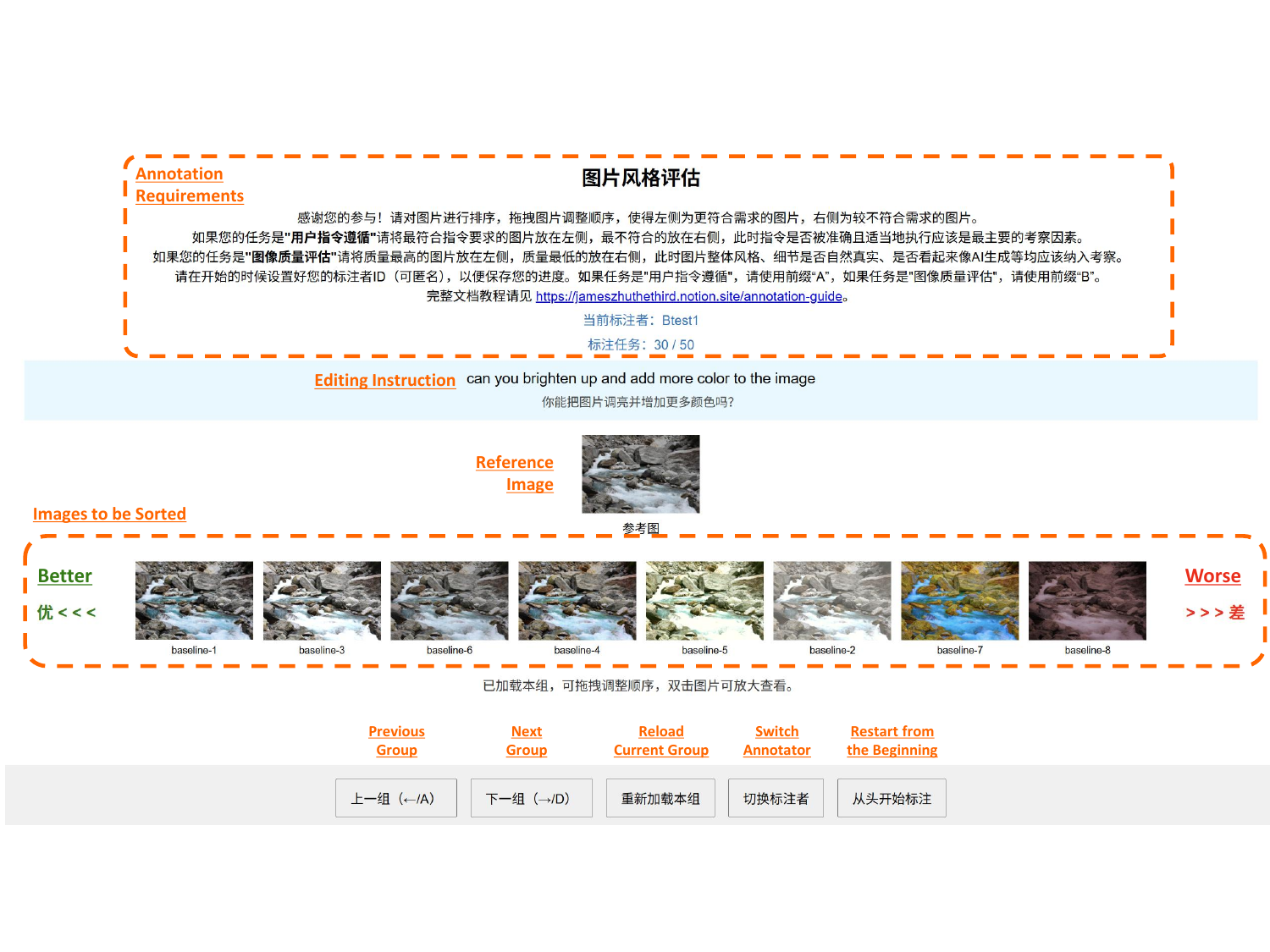} 
\caption{Screenshot of the user study interface. For a given original image and instruction (top), participants were asked to rank the edited results from different methods (bottom) by dragging and dropping them in order of preference or compliance.}
\label{fig:user_study_interface}
\end{figure}

\Cref{fig:user_study_interface} shows a screenshot of the ranking interface used in our user study, illustrating how participants evaluated the different editing outputs for a given instruction.

We report Friedman tests and Kendall's $W$ for inter-user agreement in the user study(\Cref{tab:significance}). Both tasks show high significance ($p{<}10^{-7}$). Pairwise comparisons confirm $p{<}0.05$ for all baseline pairs adjacent ranks in Task A, and all pairs except Gemini(7th) vs.\ GPT-4.1(8th) in Task B. The moderate global $W$ reflects natural variation in aesthetic preferences, while per-image $W{\approx}0.5$ indicates \textbf{reasonable agreement} when judging individual samples. 

\begin{table}[h]
\centering
\caption{Statistical significance of user study rankings.}
\label{tab:significance}
\setlength{\tabcolsep}{5pt}
\renewcommand{\arraystretch}{1}
\scriptsize
\resizebox{\linewidth}{!}{
\begin{tabular}{lcccc}
\toprule
\textbf{Task} & $\chi^2(7)$ & $p$-value & Kendall's $W$ & Per-image $W$ \\
\midrule
A: Instruction Following & 5544.46 & $<10^{-7}$ & 0.288 & $0.505{\pm}0.162$ \\
B: Image Quality & 3832.96 & $<10^{-7}$ & 0.196 & $0.544{\pm}0.169$ \\
\bottomrule
\end{tabular}
}
\end{table}

\section{Additional User Study Results}

Besides the compact user-study table in the main paper, we additionally report expanded automatic and judge-based metrics on the same 50 evaluation samples in \Cref{tab:user-study-extended}. We include two more generative baselines: \texttt{PatchDPO}~\cite{huang2025patchdpo} and GenArtist~\cite{wang2024genartist}. Besides L1/L2 distance metrics used in main paper, we also include Cosine Similarity with reference image using features extracted by CLIP~\cite{radford2021learning}, and also employ LLM-as-Judge (gemini-2.5-pro) to rate 0-10 score based on \textbf{Ins}truction Following, Reference \textbf{Sim}ilarity, and Image \textbf{Qua}lity. \agent{} shows \textbf{consistant and competitive performance} among vaiours metrics and baselines.

\clearpage

\section{Prompts Used}

\begin{tcolorbox}[breakable,colback=black!5!white, colframe=black!75!white, title=System Prompt for Image Analysis]
You are an expert image editor. You know how to edit a photo and make it look nicer. In this task, you will be given an input image, a reference output image, and a text instruction that describes the task. Your task is to determine the JSON object of parameters that were used in an image editor to transform the input image into the output image.\\

The available parameters are described in the following table:\\
\verb|{image_edit_tools}|
\end{tcolorbox}

\begin{tcolorbox}[breakable,colback=black!5!white, colframe=black!75!white, title=User Prompt for Image Analysis]
\begin{verbatim}
{instruction_template}

{operator_constraint}

{chain_of_thought}
<answer>
```json
{
    "exposure": 10,
    "sharpness": -30,
    ... // other parameters
}
```
</answer>
\end{verbatim}
\end{tcolorbox}

\begin{tcolorbox}[breakable,colback=black!5!white, colframe=black!75!white, title=Instruction templates of User Prompt for Image Analysis]
\begin{lstlisting}[breaklines=true, basicstyle=\ttfamily\small]
-1: "The user instruction is not provided. You need compare and analyze the input and output images to determine the adjustments made.",
0: "You can follow user instructions to determine the adjustments made:\n(From User) {instruction}",
1: "You can follow expert instructions to determine the adjustments made:\n(From Expert) {instruction}",
2: "You can follow user preferences to determine the adjustments made:\n{instruction}"
\end{lstlisting}
\end{tcolorbox}

\begin{tcolorbox}[breakable,colback=black!5!white, colframe=black!75!white, title=Operator constraint templates of User Prompt for Image Analysis]
\begin{lstlisting}[breaklines=true, basicstyle=\ttfamily\small]
0: "If a function or parameter is not mentioned in the instruction, it could also be included in the output. Analyze the images and instruction to determine which functions and parameters are relevant.",
1: "You MUST include and ONLY use the following operators in your output: \n{operators}\nDo not include any other operators.",
2: "During previous attempts, You are told that you MUST include and ONLY use selected functions. If you are sure that the used functions are not enough to make the image nicer, you can ignore the operator constraint and use any available functions provided above in the table."
\end{lstlisting}
\end{tcolorbox}

\begin{tcolorbox}[breakable,colback=black!5!white, colframe=black!75!white, title=Chain of thought templates of User Prompt for Image Analysis]
\begin{lstlisting}[breaklines=true, basicstyle=\ttfamily\small]
0: "Your output should be a dictionary in JSON format with the modified parameters:",
1: "Think step by step to determine the adjustments made in the image edit. Consider the input and output images, and the user instruction if provided. First, provide your thoughts about how to optimize the input image, reasoning just like your usual tasks that ONLY an input image is provided. DO NOT reveal that you have known the user instructions or the reference output image (DO NOT mention them!). DO NOT contain exact parameter numbers but use expressions of scale. You should follow the template below:\n<think> your thoughts about how to make this photo nicer </think>",
2: """Think step by step to determine the adjustments made in the image edit. Consider the input and output images, and the user instruction if provided. First, provide your thoughts about how to optimize the input image, reasoning just like your usual tasks that ONLY an input image is provided. DO NOT reveal that you have known the user instructions or the reference output image (DO NOT mention them!). DO NOT contain exact parameter numbers but use expressions of scale. You should follow the template below:
<think>
1. Brightness/Exposure: Check if the image needs to be brighter or darker overall (use brightness, exposure, highlights, shadows, whites, blacks).
2. Color: Note any possible change in color intensity or warmth (use saturation, vibrance, temperature, tint).
3. Contrast/Sharpness: See if contrast or sharpness should be changed (use contrast, sharpness, fade, grain).
4. Effects/Objects: Look for possible vignetting, fading, cropping, rotation, or object-level edits (crop, rotate, flip, inpaint_obj, rotate_obj, etc).
5. Summary: List all likely parameter changes and values that can make the input image look nicer.
</think>
""",
    3: """Think step by step to determine the adjustments made:
<think>
1. General Optimization: 
   - Brightness/Exposure: Check if the image needs to be brighter or darker
   - Color: Note any color adjustments needed
   - Contrast/Sharpness: Evaluate contrast and sharpness
   - Effects/Objects: Identify needed edits like cropping or rotation

2. User Preference Analysis:
   {preference_analysis}

3. Integrated Adjustments: Combine general improvements with user preferences
</think>
"""
\end{lstlisting}
\end{tcolorbox}

\begin{tcolorbox}[breakable,colback=black!5!white, colframe=black!75!white, title=User Prompt for Image Reflection]
\begin{verbatim}
{instruction_template}
{operator_constraint}

Parameters used in the previous best 
round of edit:
{best_params}
\end{verbatim}
You should reflect on the previous best edit and determine if any changes to parameters or functions are needed.
\begin{verbatim}
{chain_of_thought}
<answer>
```json
{
    "exposure": 10,
    "sharpness": -30,
    ... // other parameters
}
```
</answer>
\end{verbatim}
\end{tcolorbox}

\begin{tcolorbox}[breakable,colback=black!5!white, colframe=black!75!white, title=Chain of thought templates of User Prompt for Image Reflection]
\begin{lstlisting}[breaklines=true, basicstyle=\ttfamily\small]
0: "Your output should be a dictionary in JSON format with the modified parameters:",
1: "Think step by step to determine the adjustments made in the image edit. Consider the input and output images, and the user instruction if provided. First, provide your thoughts about how to optimize the input image, reasoning just like your usual tasks that ONLY an input image is provided. DO NOT reveal that you have known the user instructions or the reference output image or the previous edit image (DO NOT mention them!). DO NOT contain exact parameter numbers but use expressions of scale. You should follow the template below:\n<think> your thoughts about how to make this photo nicer </think>",
2: """Think step by step to determine the adjustments made in the image edit. Consider the input and output images, and the user instruction if provided. First, provide your thoughts about how to optimize the input image, reasoning just like your usual tasks that ONLY an input image is provided. DO NOT reveal that you have known the user instructions or the reference output image or the previous edit image (DO NOT mention them!). DO NOT contain exact parameter numbers but use expressions of scale. You should follow the template below:
<think>
1. Brightness/Exposure: Check if the image needs to be brighter or darker overall (use brightness, exposure, highlights, shadows, whites, blacks).
2. Color: Note any possible change in color intensity or warmth (use saturation, vibrance, temperature, tint).
3. Contrast/Sharpness: See if contrast or sharpness should be changed (use contrast, sharpness, fade, grain).
4. Effects/Objects: Look for possible vignetting, fading, cropping, rotation, or object-level edits (crop, rotate, flip, inpaint_obj, rotate_obj, etc).
5. Summary: List all likely parameter changes and values that can make the input image look nicer.
</think>
"""
\end{lstlisting}
\end{tcolorbox}

\begin{tcolorbox}[breakable,colback=black!5!white, colframe=black!75!white, title=System Prompt for Summary Reward]
You are an expert image editor. You know every function and parameter as well as their effect in an image editor. You will be given a user instruction and a summarized user preference. Your task is to determine whether the summarized user preference is consistent with the user instruction and report a consistent score between -10 and 10. You can follow the cases below to determine the score:\\
Reference: Make the image brighter and vivid, and add sharpness to it.\\
Prediction 1: The user prefers a brighter and more colorful image, also make the image sharper.\\
Score 1: 10\\
Prediction 2: The user prefers a brighter and less colorful image.\\
Score 2: -5\\
Prediction 3: The user prefers a image high contrast and sharpness.\\
Score 3: 3\\
Prediction 4: The user wants to make the image nicer.\\
Score 4: 0\\
Prediction 5: Today is a sunny day. (Irrelevant answer, or in a different language, or kept repeating the same sentence, or other nonsense)\\
Score 5: -10
\end{tcolorbox}

\begin{tcolorbox}[breakable,colback=black!5!white, colframe=black!75!white, title=User Prompt for Summary Reward]
Now, give your score as a single integer between -10 and 10 based on the user instruction and the summarized user preference; do not output any other text.
\begin{verbatim}
Reference: {reference}
Prediction: {prediction}
Score:
\end{verbatim}
\end{tcolorbox}

\begin{tcolorbox}[breakable,colback=black!5!white, colframe=black!75!white, title=System Prompt for Generating Instructions]
You are an expert image editor. You know every function and parameter as well as their effect in an image editor. You will be given a group of slider parameters (range -100~100) that were used to edit a photo. Your task is to convert them into a concise, natural, and specific image editing instruction, as if it were written by an amateur user.\\

The available parameters are described in the following table:\\
\verb|{image_edit_tools}|
\end{tcolorbox}

\begin{tcolorbox}[breakable,colback=black!5!white, colframe=black!75!white, title=User Prompt for Generating Instructions]
Given the following image editing parameters and related technical instruction written by an expert, generate a concise and natural image editing instruction that an amateur user might give (i.e., I want.../Make it.../Could you.../This image is too...). Do not mention numeric values or technical terms. Focus on describing the visual effect or style preferred in everyday language, using adverbs of degree where appropriate, instead of listing parameters. Do not output any other text.\\
\begin{verbatim}
Parameters:
{params}
Expert Instruction:
{expert_instruction}
Amateur User Instruction:
\end{verbatim}
\end{tcolorbox}

\begin{tcolorbox}[breakable,colback=black!5!white, colframe=black!75!white, title=System Prompt for Intent Following Judger]
You are an image editing evaluator. Evaluate how well the edited image, produced by modifying the original according to the given instruction, adheres to the specified editing requirements. You are only required to judge how closely the edit follows the instruction instead of the image quality. Score from 0 to 10:\\
* 10: Perfectly follows instruction, i.e., all changes are appropriate\\
* 7 - 9: Mostly follows with minor issues, i.e., missing minor changes, the modifications are over/under done\\
* 4 - 6: Partially follows but significant issues, i.e., some changes are inappropriate or missing major changes, or just like the original image\\
* 1 - 3: Makes opposite changes, i.e., changes are irrelevant and contradict the instruction with negative impact, the image looks obviously worse\\
* 0: Completely wrong, i.e., the image is entirely changed, unclear, or distorted in a way that contradicts the instruction\\

Please focus on the detail shift between two images, the change could be small but important.

Return only the score as an integer between 0 and 10.
\end{tcolorbox}

\begin{tcolorbox}[breakable,colback=black!5!white, colframe=black!75!white, title=User Prompt for Intent Following Judger]
\begin{verbatim}
**Original Image:** 
<original_image>
**Edited Image:** 
<edited_image>
**Editing Instruction:** 
{instruction}
\end{verbatim}

Score:
\end{tcolorbox}

\begin{tcolorbox}[breakable,colback=black!5!white, colframe=black!75!white, title=System Prompt for Target Gap Judger]
You are an image editing evaluator. Evaluate how closely the expert-provided target image aligns with the image produced by editing the original image according to the given instruction. You are not required to judge how closely the edit follows the instruction. Instead, consider the visual similarity between edited image and target image. Score from 0 to 10:\\
* 10: Edited image matches target perfectly, i.e., the edited image is nearly identical to the target image\\
* 7 - 9: Close to target with small differences, i.e., minor shift on colors or brightness, the edited image looks more similar to target than original image\\
* 4 - 6: Some similarity but major differences, i.e., significant color/brightness changes, blurs, or artifacts, or just like the original image\\
* 1 - 3: Completely different from target, i.e., wrong objects, extreme distortions, or irrelevant edits, not similar to either target or original image\\
* 0: No resemblance to target, i.e., the image is entirely changed, unclear, mosaic, or distorted in a way that contradicts the target\\

Please focus on the similarity between images, if the edited image is very close to the target image, give a high score, if the edited image is very close to the original image, give a near zero score, if the edited image is very different from the target image and the original image, give a negative score. Please focus on the detail shift between three images, the change could be small but important.

Return only the score as an integer between 0 and 10.
\end{tcolorbox}

\begin{tcolorbox}[breakable,colback=black!5!white, colframe=black!75!white, title=User Prompt for Target Gap Judger]
\begin{verbatim}
**Original Image:** 
<original_image>
**Edited Image:** 
<edited_image>
**Target Image:** 
<target_image>
**Editing Instruction:** 
{instruction}
\end{verbatim}

Score:
\end{tcolorbox}

\begin{tcolorbox}[breakable,colback=black!5!white, colframe=black!75!white, title=System Prompt for Image Quality Judger]
You are an image editing evaluator. Evaluate the quality of the image produced by editing the original image according to the given instruction. You are not required to judge how closely the edit follows the instruction. Instead, consider the overall visual style, the naturalness and realism of the details, and any indications that the image may be AI‑generated. Score from 0 to 10:\\
* 10 : Edited image is of high quality, i.e., clear, realistic, well-exposed, etc.\\
* 7 - 9: Edited image is generally good with minor flaws, i.e., slight noise, imperfect exposure, etc.\\
* 4 - 6: Edited image has noticeable quality issues, i.e., low resolution, unnatural details, weird colors, etc.\\
* 1 - 3: Edited image is of poor quality, i.e., blurry, artifacts, unrealistic, mosaic, etc.\\
* 0: Edited image is of very low quality, i.e., pure white/black, severely distorted, extremely blurry, or completely unrealistic

Please focus on the quality of the edited image itself, both general appearance and fine details.

Return only the score as an integer between 0 and 10.
\end{tcolorbox}

\begin{tcolorbox}[breakable,colback=black!5!white, colframe=black!75!white, title=User Prompt for Image Quality Judger]
\begin{verbatim}
**Original Image:** 
<original_image>
**Edited Image:** 
<edited_image>
**Editing Instruction:** 
{instruction}
\end{verbatim}

Score:
\end{tcolorbox}

\end{document}